\documentclass[12pt]{article}
\usepackage{algorithm}
\usepackage{algorithmic}
\usepackage{latexsym}
\usepackage{cite}
\usepackage{enumitem}
\usepackage{amssymb,mathrsfs}
\usepackage{booktabs}
\usepackage{subfigure}
\usepackage{float}
\usepackage{color}
\usepackage{amsmath}
\usepackage[colorlinks,
            linkcolor=blue,
            anchorcolor=red,
            citecolor=blue]{hyperref}
\usepackage{bm}
\usepackage{natbib}
\usepackage{graphicx}
\usepackage{float}
\usepackage{subfigure}
\usepackage{wrapfig}
\usepackage{amsthm}
\usepackage{epstopdf}
\usepackage{multirow}
\usepackage[table,xcdraw]{xcolor}
\usepackage{booktabs}

\setcitestyle{round}
\allowdisplaybreaks[4]

\newtheorem{theorem}{Theorem}[part]

\newtheorem{assumption}{Assumption}[part]
\newtheorem{lemma}{Lemma}[part]

\allowdisplaybreaks[4]
\begin{document}
\title{A Trace-restricted Kronecker-Factored Approximation to Natural Gradient}

\author{Kai-Xin Gao\thanks{Equal contribution. School of Mathematics, Tianjin University, Tianjin 300072, P.R. China.
Email: \{gaokaixin, liuxiaolei, huangzhenghai\}@tju.edu.cn.}
\and
Xiao-Lei Liu$^\ast$
\and
Zheng-Hai Huang$^\ast$
\and
Min Wang\thanks{Central Software Institute, Huawei Technologies, Hangzhou 310052,  P.R. China. Email: \{wangmin106, wang1, fan.yu\}@huawei.com.}  \\
\and
Zidong Wang$^\dag$    \\
\and
Dachuan Xu\thanks{Corresponding author.
Department of Operations Research and Scientific Computing, Beijing University of Technology, Beijing 100124, P.R. China.
Email: xudc@bjut.edu.cn.}\\
\and
Fan Yu$^\dag$
}

\date{}

\maketitle

\begin{abstract}
\noindent
Second-order optimization methods have the ability to accelerate convergence by modifying the gradient through the curvature matrix. There have been many attempts to use second-order optimization methods for training deep neural networks. Inspired by diagonal approximations and factored approximations such as Kronecker-Factored Approximate Curvature (KFAC), we propose a new approximation to the Fisher information matrix (FIM) called Trace-restricted Kronecker-factored Approximate Curvature (TKFAC) in this work, which can hold the certain trace relationship between the exact and the approximate FIM. In TKFAC, we decompose each block of the approximate FIM as a Kronecker product of two smaller matrices and scaled by a coefficient related to trace. We theoretically analyze TKFAC's approximation error and give an upper bound of it. We also propose a new damping technique for TKFAC on convolutional neural networks to maintain the superiority of second-order optimization methods during training. Experiments show that our method has better performance compared with several state-of-the-art algorithms on some deep network architectures.

\vspace{3mm}


\vspace{3mm}

\noindent {\bf }\hspace{2mm}

\end{abstract}

\section{Introduction}

Recently, deep learning has made great progress in a host of application areas, such as computer vision and natural language processing. However, as the size of deep neural networks (DNNs) increases rapidly, more and more computational power and time are needed to train these models. Therefore, efficient algorithm is necessary for training DNNs.

Stochastic Gradient Descent (SGD) \citep{sgd1991} and its extension Stochastic Gradient Descent with momentum (SGDM) \citep{sgdm1999} are ubiquitously used for training DNNs, due to low computational costs and eases of implementation. However, SGD is a first-order optimization method and only considers first-order gradient information, which leads to some deficiencies, including sensitivity to hyperparameter
settings and relatively-slow convergence. By using the curvature matrix to correct gradient, second-order optimization methods have the ability to solve these deficiencies efficiently. The most well-known second-order optimization method may be the Newton's method. Natural gradient descent (NGD) \citep{nat1998}, which gives the steepest descent direction in the space of distributions, also can be viewed as a second-order optimization  method \citep{new2014}. However, second-order optimization methods are clearly not computationally competitive with first-order alternatives, because they need to invert the large curvature matrix (the Hessian matrix for Newton's method and the Fisher Information Matrix (FIM) for NGD) whose dimension is the number of model's parameters. In practice, they require cubic computation time and quadratic storage for every update. Obviously, it is impractical to apply second-order optimization methods for training DNNs with hundreds of millions of parameters directly. So many approximate methods have been proposed.

Many approximation methods can be viewed as diagonal approximations of the curvature matrix (\citeauthor{diag1988}, \citeyear{diag1988}; \citeauthor{adagrad2011}, \citeyear{adagrad2011}; \citeauthor{rmsprop2012}, \citeyear{rmsprop2012};\citeauthor{adam2014}, \citeyear{adam2014}; \citeauthor{hessian2020}, \citeyear{hessian2020}), and these algorithms are computationally tractable. However, we know that the curvature matrix of DNNs' objective function is highly non-diagonal, and hence, diagonal approximations lose much curvature matrix information. More complex algorithms are not limited to diagonal approximations, but instead focus on some correlations between parameters of neural networks and use the non-diagonal part of the preconditioner matrix, such as quasi-Newton methods (\citeauthor{nqn2000}, \citeyear{nqn2000}; \citeauthor{aqn2016}, \citeyear{aqn2016}; \citeauthor{qn2019}, \citeyear{qn2019}; \citeauthor{qn2020}, \citeyear{qn2020}), Hessian-Free optimization approach (\citeauthor{hf2010}, \citeyear{hf2010}; \citeauthor{hf2011}, \citeyear{hf2011}; \citeauthor{hf2013}, \citeyear{hf2013}; \citeauthor{hf2017}, \citeyear{hf2017}), Kronecker-factored Approximate Curvature (KFAC) (\citeauthor{kfac2015}, \citeyear{kfac2015}; \citeauthor{kfc2016}, \citeyear{kfc2016}; \citeauthor{kfacr2018}, \citeyear{kfacr2018}; \citeauthor{nkfac2018}, \citeyear{nkfac2018}; \citeauthor{ekfac2018}, \citeyear{ekfac2018}; \citeauthor{nekfac2018}, \citeyear{nekfac2018}). These methods have achieved advanced performance on some complicated DNN models and training tasks.

In this paper, our main focus will be on the NGD. Motivated by the diagonal approximations, we think that the diagonal elements' information of the curvature matrix plays an important role compared to other elements, so we pay more attention the diagonal information and keep the trace equal in approximation. Inspired by both diagonal and factored approximations, we present a new approximation to the FIM of DNNs Trace-restricted Kronecker-factored Approximate Curvature (TKFAC) based on the quadratic form estimator proposed in \citet{model2019} in this work.

Our approximation is built by two steps. In the first step, the FIM is approximated to a block-diagonal matrix according to the layers of DNNs as KFAC \citep{kfac2015}. In the second step, every block matrix is decomposed into a constant multiple of the Kronecker product of two smaller matrices and keep the traces equal based on the quadratic form estimator proposed in \citet{model2019}. For DNNs, we first consider TKFAC on the fully-connected layers and give Theorem \ref{thm1} to compute the factors in TKFAC. Then, the block matrix of convolutional layers can also be decomposed efficiently under a reasonable assumption by Theorem \ref{thm3}. We also discuss the approximation effect of TKFAC and give a upper bound of its approximation error in Theorem \ref{apperr}. Next, we consider the damping used in TKFAC and provide a new damping technique TKFAC on convolutional neural networks (CNNs). Finally, to evaluate our proposed methods, we consider two variants of TKFAC compared with SGDM, Adam, and KFAC (\citeauthor{kfac2015}, \citeyear{kfac2015}; \citeauthor{kfc2016}, \citeyear{kfc2016}) on the deep auto-encoder problems using fully-connected neural networks (FNNs) and the image classification tasks using CNNs. Experimental results show that TKFAC is an effective method.


Our contributions are briefly summarized as follows:
\begin{enumerate}
  \setlength{\itemsep}{3pt}
  \setlength{\parsep}{0pt}
  \setlength{\parskip}{0pt}
  \item Motivated by both diagonal and factored approximations, a new approximation to the FIM called TKFAC is proposed based on a quadratic form estimator, in which
      the information about the sum of diagonal elements between the exact and the approximate FIM can be maintained.
  \item The approximation effect of TKFAC is discussed. The visualization results of approximation errors show that TKFAC is a indeed effective approximation to the FIM. Furthermore, an upper bound of TKFAC's approximation error is given and proved, which is less than the upper bound of KFAC's error in general cases. Experimental results on MNIST show that TKFAC can keep smaller approximation error than KFAC during training.
  \item Two damping techniques are adopted for TKFAC, including the normal damping used in previous works and the new automatic tuning damping proposed in this work, which can avoid the problem that the damping is large enough to dominate FIM and transforms the second-order optimizer into the first-order one in our experiments.
  \item Two variants of TKFAC, that are TKFAC\_nor using the normal damping and TKFAC\_new using the new damping, are compared with several state-of-the-art algorithms. TKFAC\_nor has better optimization performance while has good generalization ability. TKFAC\_new converges faster than baselines and TKFAC\_nor, and especially it has better performance on the ResNet network.
\end{enumerate}

The rest parts of the paper are organized as follows. Section \ref{sec-2} introduces some related works. Section \ref{sec-3} gives the background and notation for natural gradient and neural networks, in which we also give some important properties of the Kronecker product and introduce the quadratic form estimator. Section \ref{sec-4} describes our approximation to the FIM of fully-connected layers, convolutional layers and gives some discussion of the approximation error. Section \ref{sec-5} introduces some tricks we adopted for TKFAC and gives a new technique scheme for TKFAC on CNNs. Section \ref{sec-6} describes our experimental setup and gives the evaluation results of TKFAC. Section \ref{sec-7} gives the conclusion.
\section{Related work}\label{sec-2}

There have been many attempts to apply NGD or its approximations to train DNNs. The main computational challenge to use NGD is to store and invert the FIM. Recently, some works have considered the efficient Kronecker-factored approximation to the FIM, such as KFAC (\citeauthor{kfac2015}, \citeyear{kfac2015}; \citeauthor{kfc2016}, \citeyear{kfc2016}; \citeauthor{kfacr2018}, \citeyear{kfacr2018},\citeauthor{kfac2015}, \citeyear{kfac2015}; \citeauthor{nkfac2018}, \citeyear{nkfac2018}; \citeauthor{nekfac2018}, \citeyear{nekfac2018}) and Eigenvalue-corrected Kronecker Factorization (EKFAC) \citep{ekfac2018}.

KFAC was first proposed in \citet{kfac2015} for FNNs, which provides a successful approximate natural gradient optimizer. KFAC starts with a block-diagonal approximation of the FIM (with blocks corresponding to entire layers), and then approximates each block as a Kronecker product of two much smaller matrices. By the property that the inverse of a Kronecker product of two matrices is equal to the Kronecker product of their inverses, each block can be inverted tractably. Furthermore, the inverse matrix of the exact FIM can be computed efficiently. Therefore, NGD can be used to train DNNs efficiently. Then, KFAC is extended to convolutional neural networks \citep{kfc2016}, recurrent neural networks \citep{kfacr2018} and variational Bayesian neural networks \citep{nkfac2018,nekfac2018} and shows significant speedups during training. \citet{ekfac2018} tracks the diagonal variance in the Kronecker-factored eigenbasis and proposes EKFAC based on KFAC. What's more, KFAC also has been applied to large-scale distributed computing for DNNs and shows excellent experimental performance (\citeauthor{dis2017}, \citeyear{dis2017}; \citeauthor{lar2019}, \citeyear{lar2019}; \citeauthor{dis2020}, \citeyear{dis2020}). 

Covariance matrices are of great importance in many fields and there have been many new methodological approaches to covariance estimation in the large dimensional case (i.e., the dimension of the covariance matrix is large compared with the
sample size) (\citeauthor{mat2004}, \citeyear{mat2004}; \citeauthor{mat2011}, \citeyear{mat2011}; \citeauthor{mat2016}, \citeyear{mat2016}; \citeauthor{model2019}, \citeyear{model2019}). Recently, \citet{model2019} proposed an estimator of the Kronecker product model of the covariance matrix called quadratic form estimator and shows that this estimator has good properties in theory. In this work, we draw inspiration from the Kronecker-factored approximation to the FIM and the quadratic form estimator to covariance matrix and propose TKFAC.
\section{Background and Notation} \label{sec-3}

\subsection{Deep neural networks}
Given a training dataset $D_{train}=(x,y)$ containing (input, target) examples $(x, y)$. Define $f(x,\omega)$ to be the neural network function related to input $x$, where $\omega$ are the parameters. Consider the loss function $\mathcal{L}(y,f(x,\omega))=-\log p(y|f(x,\omega))$, where $p$ represents the density function of model's predictive distribution $P$. Therefore, the loss function of a single training case can be given by $g(y,h(x,\omega))=-\log p(y|f(x,\omega))$. The objective function which we wish to minimize during training is the expected loss
$$h(\omega)=\mathbb{E}_{(x,y)\in D_{train}}[\mathcal{L}(y,f(x,\omega))]=\mathbb{E}_{(x,y)\in D_{train}}[-\log p(y|f(x,\omega))].$$
For simpllicity, throughout the rest of this paper we will use the following notation for derivatives of the loss w.r.t. some arbitrary variable $V$:
$$\mathcal{D}V = -\frac{{\rm d}\mathcal{L}(y,f(x,\omega))}{{\rm d} V}=-\frac{{\rm d}\log p(y|x,\omega)}{{\rm d} V},\quad g_l=\mathcal{D}s_l.$$

Consider a fully-connected layer of a feed-forward DNN. The computation performed in this layer can be given as follows:
$$s_l=W_la_{l-1},\quad a_l=\varphi_l(s_l)$$
where $a_{l-1}\in \mathbb{R}^{m_{l-1}}$ is the input of this layer (the activation from the previous layer), $W_l\in\mathbb{R}^{m_{l}\times m_{l-1}}$ (the bias is ignored for convenience) is the weight matrix, $\varphi_l$ is the activation function, $l$ represents the $l$-th layer and we refer to the $s_l$ as the pre-activation of this layer. By the chain rule, the derivatives of the weights are given by $\mathcal{D}W_l=g_la_{l-1}^\top$.

For a convolutional layer, the pre-activation $\mathcal{S}_l\in \mathbb{R}^{n_{l}\times o_{l-1}}$ \footnote{For convenience, we directly give the operation expressed in matrix form for convolutional layers.}is calculated as
$$\widehat{\mathcal{A}_{l-1}}={\rm im2col}(\mathcal{A}_{l-1})\in \mathbb{R}^{n_{l-1}k_l^2\times o_{l-1}},\quad
\mathcal{S}_{l}=W_l \widehat{\mathcal{A}_{l-1}}\in \mathbb{R}^{n_{l}\times o_{l-1}}.$$
where ${\rm im2col}(\cdot)$ is a function to rearrange image blocks into columns, $\mathcal{A}_{l-1}\in \mathbb{R}^{n_{l-1}\times o_{l-1}}$ is the input of this layer, $W_l\in \mathbb{R}^{n_{l}\times n_{l-1}k_l^2}$ (the bias is ignored for convenience) is the weight matrix, $n_{l-1}$ and $n_l$ are the numbers of input and output channels, $o_{l-1}$ is the number of spatial locations and $k_l$ is the kernel size. The gradient of the weights is computed as $\mathcal{D}W_l=\mathcal{D}\mathcal{S}_l(\widehat{\mathcal{A}_{l-1}})^\top\in\mathbb{R}^{n_l\times n_{l-1}k_l^2}$.

We will define $\omega=[\omega_1,\omega_2,\cdots,\omega_L]=[{\rm vec}(W_1)^\top,{\rm vec}(W_2)^\top,\cdots,{\rm vec}(W_L)^\top]^\top$ in the rest of this paper, which is a vector consists all the parameters of a DNN with $L$ layers and $\omega_l$ consists all the parameters of the $l$-th layer.
\subsection{Natural gradient}

NGD can be interpreted as a second-order optimization method \citep{new2014}, in which the curvature matrix is the FIM, given by
$$F=\mathbb{E}[\mathcal{D}\omega\mathcal{D}\omega^\top]={\rm cov}(\mathcal{D}\omega,\mathcal{D}\omega),$$
where the expectation is associated with $x$ sampled from the training distribution and $y$ sampled according to the prediction distribution $P$.

The natural gradient is usually defined as $F^{-1}\nabla_{\omega}h$ \citep{nat1998}, which provides the update direction for natural gradient descent. Then the parameters are updated by
$$\omega\leftarrow\omega-\alpha F^{-1}\nabla_{\omega}h,$$
where $\alpha$ is learning rate and $\nabla_{\omega}h$ is the gradient. The natural gradient gives the steepest descent direction in the distribution space, as measured by the KL-divergence \citep{geo2000}. This is in contrast to the standard gradient, which gives the fastest change direction in the parameter space under the Euclidean metric. More discussion of the natural gradient can refer to \citet{new2014}.

\subsection{Kronceker product}
In this paper, we will use many convenient properties of the Kronceker product and we will introduce them in this subsection. The related concepts and properties mainly come from \citet{kro1986} and \citet{kro1997}.

For an $m\times n$ matrix $A=(a_{ij})$ and a $p\times q$ matrix $B=(b_{ij})$, the Kronceker product is denoted by the symbol $A\otimes B$ and is defined to be the $mp\times nq$ matrix
\begin{equation*}
  \left(
  \begin{array}{ccc}
    a_{11}B & \cdots & a_{1n}B\\
    \vdots & \ddots & \vdots\\
    a_{m1}B & \ldots & a_{mn}B
  \end{array}
\right).
\end{equation*}
Some useful properties of the Kronceker product for matrices $A,B,C,D$ and vectors $u,v$ are summarized as follows:
\begin{itemize}
  \setlength{\itemsep}{3pt}
  \setlength{\parsep}{0pt}
  \setlength{\parskip}{0pt}
  \item $(A\otimes B)^\top=A^\top \otimes B^\top$
  \item $(A\otimes B)^{-1}=A^{-1} \otimes B^{-1}$
  \item $(A\otimes B)(C\otimes D)=(AC\otimes BD)$
  \item tr($A\otimes B$)=tr($A$)$\otimes$tr($B$)
  \item $\|A\otimes B\|_F=\|A\|_F \otimes \|B\|_F$, where $\|\cdot\|_F$ is the Frobenius norm
  \item $(u\otimes v)$=vec($vu^\top$)
  \item vec($ABC$)=($C^\top\otimes A$)vec($B$)
\end{itemize}


In addition to these basic properties of the Kronecker product, we also need the following lemma, which is the Lemma 4 in \citet{kro1986}.

\begin{lemma}\label{k-9}
For an $m\times n$ matrix $A=(a_{ij})$ and a $p\times q$ matrix $B=(b_{ij})$, $K_{mn}$ is the matrix satisfying vec($A^\top$)=$K_{mn}$vec($A$), which is only depends on $m,n$ but not on the values $a_{11},a_{12},\cdots, a_{mn}$. Then,
$$K_{pm}(A\otimes B)=(B\otimes A)K_{qn}, \quad K_{pm}(A\otimes B)K_{qn}=B\otimes A.$$
\end{lemma}

\subsection{Quadratic form estimator}

\citet{model2019} considered the following Kronecker product mode. Consider a covariance matrix $\Theta\in\mathbb{R}^{n\times n}$. Let $n=n_1\times n_2\times\cdots \times n_r$, where $n_j\in \mathbb{Z}$ and $n_j\geq 2$ for $j = 1, \cdots, r$. Suppose that
\begin{equation}\label{model}
\Theta=\delta\times \Theta_1\otimes\Theta_2\cdots\otimes\Theta_{r}
\end{equation}
where $0<\delta<\infty$ is a scalar parameter and $\Theta_j\in \mathbb{R}^{n_j\times n_j}$ satisfying tr($\Theta_j$)=$n_j$ for $j = 1, \cdots, r$.

The quadratic form estimator is introduced to identify the factors in Eq. (\ref{model}). First, $\delta$ is easily obtained by
$${\rm tr}(\Theta)=\delta\times{\rm tr}(\Theta_1)\times\cdots\times{\rm tr}(\Theta_r)=\delta n,$$
whence we have $\delta={\rm tr}(\Theta)/ n$. Other factors can be identified based on the following partial trace operator \citep{ptr2018}.

For any matrix $A\in\mathbb{R}^{mq\times nq}$, which contains $mn$ blocks $A_{ij}$ for $i\in\{1, 2, \ldots, m\}$ and $j\in\{1, 2, \ldots, n\}$ and each block $A_{ij}$ is a matrix of $q\times q$, the partial trace operator ${\rm{PTr}}(A): \mathbb{R}^{mq\times nq}\rightarrow \mathbb{R}^{m\times n}$ is defined by
$${(\rm{PTr}}(A))_{ij}={\rm{tr}}(A_{ij}),\quad \forall i\in\{1, 2, \cdots, m\}, \forall j\in\{1, 2, \cdots, n\}.$$

Define
\begin{equation*}
\chi_j=
	\left\{
	\begin{array}{ll}
	\mbox{$\Theta_2\otimes\cdots\otimes\Theta_r$} & \mbox{$j=1$},\\
	\mbox{$\Theta_{j+1}\otimes\cdots\otimes\Theta_r\otimes\Theta_1\otimes\cdots\otimes\Theta_{j-1}$} & \mbox{$j=2,\cdots,r-1$},\\
	\mbox{$\Theta_1\otimes\cdots\otimes\Theta_{r-1}$}  & \mbox{$j=r$},\\
	\end{array}
	\right.
	\end{equation*}
and the $n_j\times n_j$ matrix $\Xi_j=\delta {\rm tr}(\chi_j)\times \Theta_j$. Then we have $\Theta_j=\Xi_j/({\rm tr}(\Xi_j)/n_j)$ for $j\in\{1, 2, \cdots, r\}$. More details can be found in \citep{model2019}.

\section{Our Method} \label{sec-4}
\setcounter{equation}{0}
\setcounter{lemma}{0}
Consider a DNN with $L$ layers, the FIM can be expressed as
\begin{equation*}
  F=\mathbb{E}[\mathcal{D}\omega\mathcal{D}\omega^\top]=\left(
  \begin{array}{ccc}
    \mathbb{E}[{\rm vec}(\mathcal{D}W_1){\rm vec}(\mathcal{D}W_1)^\top] & \cdots &\mathbb{E}[{\rm vec}(\mathcal{D}W_1){\rm vec}(\mathcal{D}W_L)^\top]\\
    \vdots & \ddots & \vdots\\
    \mathbb{E}[{\rm vec}(\mathcal{D}W_L){\rm vec}(\mathcal{D}W_1)^\top] & \ldots & \mathbb{E}[{\rm vec}(\mathcal{D}W_L){\rm vec}(\mathcal{D}W_L)^\top]
  \end{array}
\right).
\end{equation*}
Our method is started with a block-diagonal approximation to the FIM, that is
\begin{align}\label{diag}
\begin{aligned}
  F={\rm diag}(F_1, F_2, \cdots, F_L),\\
\end{aligned}
\end{align}
where $F_l=\mathbb{E}[{\rm vec}(\mathcal{D}W_l){\rm vec}(\mathcal{D}W_l)^\top]\in \mathbb{R}^{m_{l-1}m_l\times m_{l-1}m_l}$ for any $l\in\{1,2,\cdots,L\}$. Although the FIM can be approximated to $L$ block matrices in Eq. (\ref{diag}), the dimension of each block matrix $F_l$ is still too large to invert $F_l$ easily in practice, so further approximation is necessary.

\subsection{TKFAC for fully-connected layers}

For a fully-connected layer, the block matrix $F_l\in \mathbb{R}^{m_{l-1}m_l\times m_{l-1}m_l}$ can be computed as
\begin{align}\label{fc1}
\begin{aligned}
F_l
&=\mathbb{E}[{\rm vec}(\mathcal{D}W_l){\rm vec}(\mathcal{D}W_l)^\top]=\mathbb{E}[{\rm vec}(g_l a_{l-1}^\top){\rm vec}(g_l a_{l-1}^\top)^\top]\\
&=\mathbb{E}[(a_{l-1} \otimes g_l)(a_{l-1} \otimes g_l)^\top]
=\mathbb{E}[(a_{l-1}a_{l-1}^\top)\otimes(g_lg_l^\top)]=\mathbb{E}[\Lambda_{l-1}\otimes \Gamma_l],
\end{aligned}
\end{align}
where $\Lambda_{l-1}=a_{l-1}a_{l-1}^\top\in \mathbb{R}^{m_{l-1}\times m_{l-1}}$ and $\Gamma_l=g_lg_l^\top\in \mathbb{R}^{m_l\times m_l}$. Then we give our decomposition of $F_l$ by Theorem \ref{thm1}.

\begin{theorem}\label{thm1}
Let $F_{l}=\mathbb{E}[\Lambda_{l-1}\otimes \Omega_l]\in \mathbb{R}^{m_{l-1}m_l\times m_{l-1}m_l}$ and suppose that $F_l$ can be decomposed as a Kronecker product scaled by a coefficient $\delta_l$, i.e.,
\begin{equation}\label{mod}
  F_{l}=\delta_l\Phi_l\otimes\Psi_l.
\end{equation}
Then, we have
\begin{eqnarray}
  \delta_l &=& \frac{\mathbb{E}[{\rm{tr}}(\Lambda_{l-1}){\rm{tr}}(\Gamma_{l})]}{{\rm tr}(\Phi_l){\rm tr}(\Psi_l)}, \label{th1}\\
  \Phi_l &=& \frac{{\rm tr}(\Phi_l)\mathbb{E}[{\rm{tr}}(\Gamma_{l})\Lambda_{l-1}]}{\mathbb{E}[{\rm{tr}}
     (\Lambda_{l-1}){\rm{tr}}(\Gamma_{l})]}\in \mathbb{R}^{m_{l-1}\times m_{l-1}}, \label{th2}\\
  \Psi_l &=& \frac{{\rm tr}(\Psi_l)\mathbb{E}[{\rm{tr}}(\Lambda_{l-1})\Gamma_{l}]}{\mathbb{E}[{\rm{tr}}(\Lambda_{l-1})
    {\rm{tr}}(\Gamma_{l})]}\in \mathbb{R}^{m_{l}\times m_{l}}. \label{th3}
\end{eqnarray}
\end{theorem}

To prove Theorem \ref{thm1}, we need to introduce the following lemma.
\begin{lemma} \label{lem1} For the two matrices $\Lambda_{l-1}\in \mathbb{R}^{m_{l-1}\times m_{l-1}}$ and $\Gamma_l\in \mathbb{R}^{m_l\times m_l}$, we have
$$
  {\rm{PTr}}(F_l)={\rm{PTr}}(\mathbb{E}[\Lambda_{l-1}\otimes\Gamma_{l}]) =\mathbb{E}[{\rm{tr}}(\Gamma_{l})\Lambda_{l-1}].
$$
\end{lemma}

\begin{proof}
For simplicity, we denote $\Lambda_{l-1}:=(\bar{a}_{ij})\in\mathbb{R}^{m\times m}$. Then,
\begin{align*}
\Lambda_{l-1}\otimes \Gamma_l=\left(
  \begin{array}{cccc}
    \bar a_{11}\Gamma_l & \ldots & \bar a_{1m}\Gamma_l\\
    \vdots & \ddots & \vdots\\
     \bar a_{m1}\Gamma_l & \ldots & \bar a_{mm}\Gamma_l\\
  \end{array}
\right).
\notag
\end{align*}
So,
\begin{align*}
\begin{aligned}
{\rm{PTr}}(F_l)&={\rm{PTr}}(\mathbb{E}[\Lambda_{l-1}\otimes \Gamma_l])\\
&={\rm{PTr}}\left(
  \begin{array}{cccc}
    \mathbb{E}[\bar{a}_{11}\Gamma_l] & \ldots &  \mathbb{E}[\bar{a}_{1m}\Gamma_l]\\
    \vdots & \ddots & \vdots\\
     \mathbb{E}[\bar{a}_{m1}\Gamma_l] & \ldots & \mathbb{E}[\bar{a}_{mm}\Gamma_l]\\
  \end{array}
\right)
=\left(
  \begin{array}{cccc}
    {\rm{tr}}(\mathbb{E}[\bar{a}_{11}\Gamma_l]) & \ldots &  {\rm{tr}}(\mathbb{E}[\bar{a}_{1m}\Gamma_l])\\
    \vdots & \ddots & \vdots\\
     {\rm{tr}}(\mathbb{E}[\bar{a}_{m1}\Gamma_l]) & \ldots & {\rm{tr}}(\mathbb{E}[\bar{a}_{mm}\Gamma_l])\\
  \end{array}
\right)\\
&=\mathbb{E}\left(
  \begin{array}{cccc}
    {\rm{tr}}(\bar{a}_{11}\Gamma_l) & \ldots &  {\rm{tr}}(\bar{a}_{1m}\Gamma_l)\\
    \vdots & \ddots & \vdots\\
     {\rm{tr}}(\bar{a}_{m1}\Gamma_l) & \ldots & {\rm{tr}}(\bar{a}_{mm}\Gamma_l)\\
  \end{array}
\right)
=\mathbb{E}[{\rm{tr}}(\Gamma_{l})\Lambda_{l-1}],\notag
\end{aligned}
\end{align*}
where the the third equality uses the definition of the partial trace operator, the forth equality uses the fact that expectation and trace can be exchanged because they are linear operations.
\end{proof}

Now, we give the proof of Theorem \ref{thm1}, using some properties of Kronecker product given in Section \ref{sec-3}.
\begin{proof} First, we show that (\ref{th1}) holds. On one hand, by (\ref{mod}) we have
\begin{equation}\label{d1}
{\rm{tr}}(F_l)=\delta_l{\rm{tr}}(\Phi_l\otimes\Psi_l)
=\delta_l{\rm{tr}}(\Phi_l){\rm{tr}}(\Psi_l),
\end{equation}
On the other hand, by (\ref{fc1}) we have
\begin{equation}\label{d2}
{\rm{tr}}(F_l)={\rm{tr}}(\mathbb{E}[\Lambda_{l-1}\otimes\Gamma_{l}])=\mathbb{E}[{\rm{tr}}(\Lambda_{l-1}\otimes\Gamma_{l})]=
  \mathbb{E}[{\rm{tr}}(\Lambda_{l-1}){\rm{tr}}(\Gamma_{l})],
\end{equation}
where the second equality follows from the fact that expectation and trace can be exchanged because they are linear operations. Thus, by (\ref{d1}) and (\ref{d2}), we obtain that
$$
\delta_l=\frac{{\rm{tr}}(F_l)}{{\rm{tr}}(\Phi_l){\rm{tr}}(\Psi_l)}
=\frac{\mathbb{E}[{\rm{tr}}(\Lambda_{l-1}){\rm{tr}}(\Gamma_{l})]}{{\rm{tr}}(\Phi_l){\rm{tr}}(\Psi_l)}.
$$

Second, we show that (\ref{th2}) holds.  On one hand, by Lemma \ref{lem1}, we have
$$
{\rm{PTr}}(F_l)=\mathbb{E}[{\rm{tr}}(\Gamma_{l})\Lambda_{l-1}].
$$
On the other hand, by using the partial trace operator and (\ref{mod}), we have
$$
{\rm{PTr}}(F_l)={\rm{PTr}}(\delta_l\Phi_l\otimes\Psi_l)=\delta_l{\rm{tr}}(\Psi_l)\Phi_l.
$$
Thus,
$$
\Phi_l=\frac{\mathbb{E}[{\rm{tr}}(\Gamma_{l})\Lambda_{l-1}]}{\delta_l{\rm{tr}}(\Psi_l)}
=\frac{{\rm{tr}}(\Phi_l)\mathbb{E}[{\rm{tr}}(\Gamma_{l})\Lambda_{l-1}]}{\mathbb{E}[{\rm{tr}}(\Lambda_{l-1}){\rm{tr}}(\Gamma_{l})]},
$$
where the second equality follows from (\ref{th1}).

Third, we show that (\ref{th3}) holds.  By Lemma \ref{k-9}, we have
$$
K_{nm}(\Lambda_{l-1}\otimes \Gamma_{l})K_{mn}=\Gamma_{l}\otimes \Lambda_{l-1}.
$$
So, by (\ref{fc1}) and (\ref{mod}), we can obtain
$$
K_{nm}F_lK_{mn}=K_{nm}(\mathbb{E}[\Lambda_{l-1}\otimes\Gamma_{l}])K_{mn}=\mathbb{E}[K_{nm}(\Lambda_{l-1}\otimes\Gamma_{l})K_{mn}]=
  \mathbb{E}[\Gamma_l\otimes\Lambda_{l-1}]
$$
and
$$
K_{nm}F_lK_{mn}=K_{nm}(\delta_l\Phi_l\otimes\Psi_l)K_{mn}=\delta_l\Psi_l\otimes\Phi_l,
$$
which lead to
$$
\mathbb{E}[\Gamma_l\otimes\Lambda_{l-1}]=\delta_l\Psi_l\otimes\Phi_l.
$$
Furthermore, by Lemma \ref{lem1} and the partial trace operator, we can obtain that
$$
\mathbb{E}[{\rm{tr}}(\Lambda_{l-1})\Gamma_l]
={\rm{PTr}}(\mathbb{E}[\Gamma_l\otimes\Lambda_{l-1}])
={\rm{PTr}}(\delta_l\Psi_l\otimes\Phi_l)
=\delta_l{\rm{tr}}(\Phi_l)\Psi_l.
$$
That is,
$$
\Psi_l=\frac{\mathbb{E}[{\rm{tr}}(\Lambda_{l-1})\Gamma_l]}{\delta_l{\rm{tr}}(\Phi_l)}
=\frac{{\rm{tr}}(\Psi_l)\mathbb{E}[{\rm{tr}}(\Lambda_{l-1})\Gamma_{l}]}{\mathbb{E}[{\rm{tr}}(\Lambda_{l-1}){\rm{tr}}(\Gamma_{l})]}.
$$

The proof is complete.
\end{proof}

Note that we simplify the model as shown by Eq. (\ref{model}) based on the $F_l$ expression in Eq. (\ref{fc1}) for the fully-connected layers. In Theorem \ref{thm1}, ${\rm{tr}}(\Phi_l)$ and ${\rm{tr}}(\Psi_l)$ are unknown, but it doesn't affect the computation in practice because we can assume that ${\rm{tr}}(\Phi_l)$ and ${\rm{tr}}(\Psi_l)$ are arbitrary constants. In practical, we may assume that ${\rm{tr}}(\Phi_l)={\rm{tr}}(\Psi_l)=1$ to reduce computing costs. In this case, (\ref{th1})-(\ref{th3}) can be simplified as
\begin{equation}\label{sim}
  \delta_l=\mathbb{E}[{\rm{tr}}(\Lambda_{l-1}){\rm{tr}}(\Gamma_{l})],\quad
  \Phi_l=\frac{\mathbb{E}[{\rm{tr}}(\Gamma_{l})\Lambda_{l-1}]}{\mathbb{E}[{\rm{tr}}(\Lambda_{l-1}){\rm{tr}}(\Gamma_{l})]},\quad
  \Psi_l=\frac{\mathbb{E}[{\rm{tr}}(\Lambda_{l-1})\Gamma_{l}]}{\mathbb{E}[{\rm{tr}}(\Lambda_{l-1}){\rm{tr}}(\Gamma_{l})]}.
\end{equation}
In the rest of this paper, we all use the simplified formulas as (\ref{sim}).

\subsection{TKFAC for convolutional layers}
For a convolutional layer, the block matrix $F_l\in\mathbb{R}^{n_{l-1}n_{l}k_l^2\times n_{l-1}n_{l}k_l^2}$ can be computed as
\begin{equation}\label{con1}
  F_l=\mathbb{E}[{\rm vec}(\mathcal{D}W_l){\rm vec}(\mathcal{D}W_l)^\top]
  =\mathbb{E}[{\rm vec}(\mathcal{D}\mathcal{S}_l(\widehat{\mathcal{A}_{l-1}})^\top){\rm vec}(\mathcal{D}\mathcal{S}_l(\widehat{\mathcal{A}_{l-1}})^\top)^\top].
\end{equation}

Note that we can't decompose Eq. (\ref{con1}) as fully-connected layers. In order to extend KFAC to convolutional layers, \citet{kfc2016} adopted some assumptions. Based on these assumptions, we propose an assumption for convolutional layers:
\begin{assumption}\label{assume}
Consider the products of activations $\mathcal{A}_{l-1}$ and pre-activation derivatives $\mathcal{D}\mathcal{S}_l$, then these products are uncorrelated at any two different spatial locations.
\end{assumption}
The rationality analysis of this assumption is given in Appendix \ref{sec-b}. Under this assumption, we can obtain that Eq. (\ref{con1}) is the sum of several Kronecker products of two smaller matrices in Theorem \ref{thm2}. Then the decomposition of $F_l$ for convolutional layers is given in Theorem \ref{thm3}.

\begin{theorem}\label{thm2}
For the block matrix $F_l$ of a convolutional layers in (\ref{con1}), let
$$
\mathcal{D}\mathcal{S}_l=[\breve{u}_1, \breve{u}_2, \cdots, \breve{u}_{o_{l-1}}]\in\mathbb{R}^{n_l\times o_{l-1}},\quad
(\widehat{\mathcal{A}_{l-1}})^\top=[\breve{a}_1, \breve{a}_2, \cdots, \breve{a}_{o_{l-1}}]^\top\in\mathbb{R}^{o_{l-1}\times n_{l-1}k_l^2},
$$
where $\breve{u}$ is the vector of each column for $\mathcal{D}\mathcal{S}_l$ and $\breve{a}$ is the vector of each column for $\mathcal{A}_{l-1}$. If the assumption \ref{assume} holds, then we have
\begin{equation}\label{th21}
   \mathcal{F}_l=\sum_{i=1}^{o_{l-1}}\mathbb{E}[\breve{a}_i\breve{a}_i^\top\otimes \breve{u}_i\breve{u}_i^\top].
\end{equation}
\end{theorem}
\begin{proof} According to (\ref{con1}) and the the assumption, we have
\begin{align*}
  \mathcal{F}_l &=\mathbb{E}[{\rm vec}(\mathcal{D}\mathcal{U}_l^\top\langle\bar{\mathcal{A}}_{l-1}\rangle)
  {\rm vec}(\mathcal{D}\mathcal{U}_l^\top\langle\bar{\mathcal{A}}_{l-1}\rangle)^\top]\\
   &=\mathbb{E}[\sum_{i=1}^{o_{l-1}}(\breve{a}_i\otimes\breve{u}_i)
   \sum_{i=1}^{o_{l-1}}(\breve{a}_i\otimes\breve{u}_i)^\top]\\
   &=\mathbb{E}[\sum_{i=1}^{o_{l-1}}(\breve{a}_i\otimes\breve{u}_i)(\breve{a}_i\otimes\breve{u}_i)^\top]\\
   &=\mathbb{E}[\sum_{i=1}^{o_{l-1}}\breve{a}_i\breve{a}_i^\top\otimes \breve{u}_i\breve{u}_i^\top]\\
   &=\sum_{i=1}^{o_{l-1}}\mathbb{E}[\breve{a}_i\breve{a}_i^\top\otimes \breve{u}_i\breve{u}_i^\top],
\end{align*}
where the second equality can be easily obtained according to the definition of Kronecker product and the vec operator, the third equality uses the assumption, and the forth equality is obtained by the property of Kronecker product.
\end{proof}

\begin{theorem}\label{thm3}\footnote{This theorem is similar to Theorem \ref{thm1}. For simplicity, we only give the the simplified formulas (let ${\rm{tr}}(\Phi_l)={\rm{tr}}(\Psi_l)=1)$.}
Let $F_l\in\mathbb{R}^{n_{l-1}n_{l}k_l^2\times n_{l-1}n_{l}k_l^2}$ be the FIM of a convolutional layer and suppose that $F_l$ can be decomposed as a Kronecker product scaled by a coefficient $\delta_l$, i.e.,
\begin{equation}\label{mod2}
  F_{l}=\delta_l\Phi_l\otimes\Psi_l,
\end{equation}
where ${\rm{tr}}(\Phi_l)={\rm{tr}}(\Psi_l)=1$. Then, we have
\begin{eqnarray}
  \delta_l
  &=&
  \sum_{i=1}^{o_{l-1}}\mathbb{E}[{\rm{tr}}(\breve{a}_i\breve{a}_i^\top){\rm{tr}}(\breve{u}_i\breve{u}_i^\top)],\nonumber\\
  \Phi_l
  &=&
  \sum_{i=1}^{o_{l-1}}\frac{\mathbb{E}[{\rm{tr}}(\breve{u}_i\breve{u}_i^\top)\times(\breve{a}_i\breve{a}_i^\top)]}
  {\mathbb{E}[{\rm{tr}}(\breve{a}_i\breve{a}_i^\top){\rm{tr}}(\breve{u}_i\breve{u}_i^\top)]}\in \mathbb{R}^{n_{l-1}k_l^2\times n_{l-1}k_l^2}, \label{th2-3}\\
  \Psi_l
  &=&
  \sum_{i=1}^{o_{l-1}}\frac{\mathbb{E}[{\rm{tr}}(\breve{a}_i\breve{a}_i^\top)\times(\breve{u}_i\breve{u}_i^\top)]}
  {\mathbb{E}[{\rm{tr}}(\breve{a}_i\breve{a}_i^\top){\rm{tr}}(\breve{u}_i\breve{u}_i^\top)]}\in
  \mathbb{R}^{n_{l}\times n_{l}}.\nonumber
\end{eqnarray}
\end{theorem}

\begin{proof}
According to the result of Theorem \ref{thm2}, this theorem can be easily proved in a similar way as the proof of Theorem \ref{thm1}.
\end{proof}

\subsection{Analysis of the approximation error}

\begin{figure}[H]
	\centering
	\vspace{-0.35cm}
	\subfigtopskip=2pt
	\subfigbottomskip=2pt
	\subfigcapskip=2pt
	\subfigure{\includegraphics[width=0.2\linewidth]{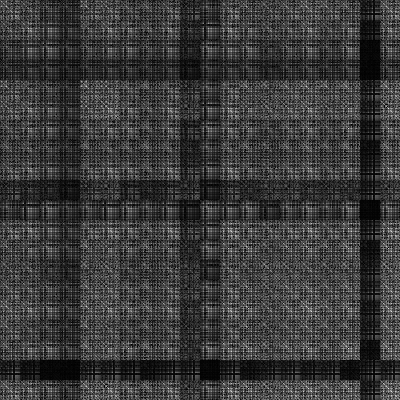}}
    \subfigure{\includegraphics[width=0.2\linewidth]{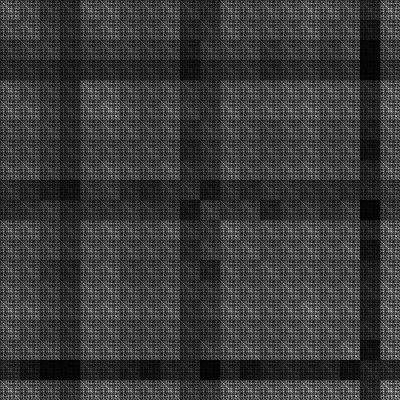}}
    \subfigure{\includegraphics[width=0.2\linewidth]{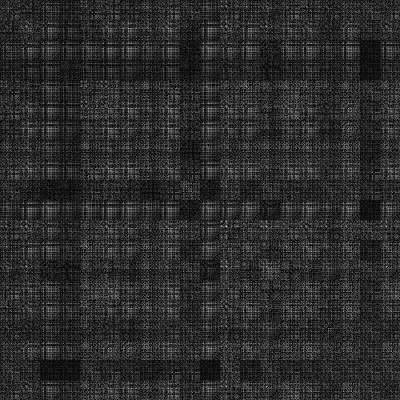}}\\
    \subfigure{\includegraphics[width=0.2\linewidth]{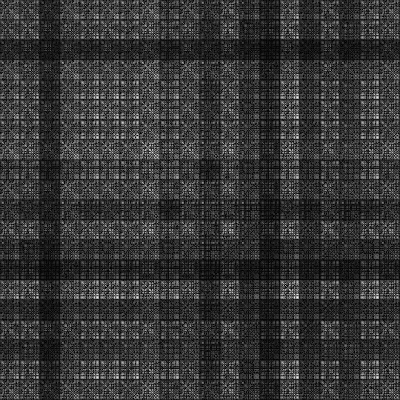}}
    \subfigure{\includegraphics[width=0.2\linewidth]{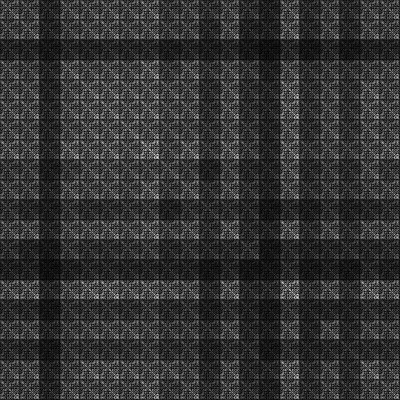}}
    \subfigure{\includegraphics[width=0.2\linewidth]{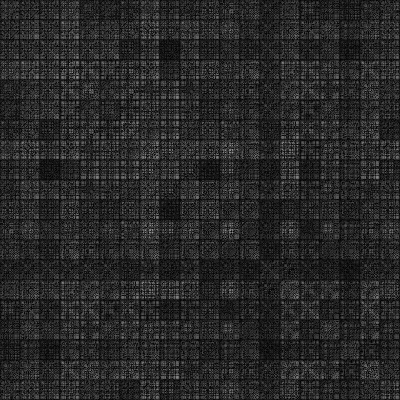}}\\
    \subfigure{\includegraphics[width=0.2\linewidth]{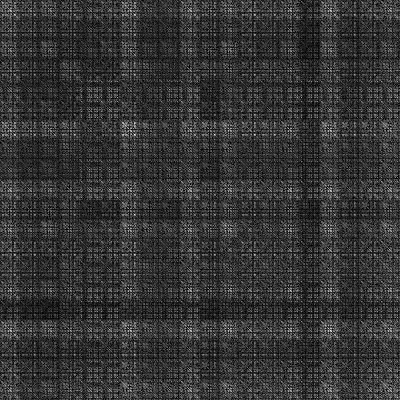}}
    \subfigure{\includegraphics[width=0.2\linewidth]{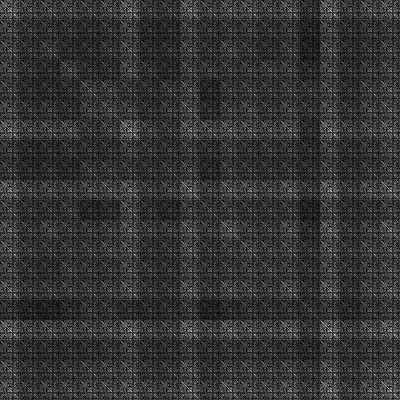}}
    \subfigure{\includegraphics[width=0.2\linewidth]{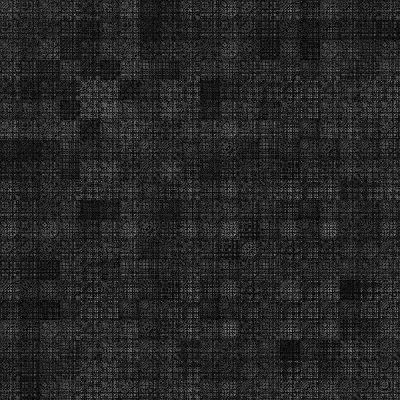}}
\caption{A comparison of the exact FIM $F$ and our approximation $F_{\rm TKFAC}$. We use a DNN to train MNIST by TKFAC. The network architecture is 196-20-20-20-20-10. We show the results of the FIM of the three layers with 20 units, which are $400\times 400$ matrices. The three figures in each row correspond to a layer. On the left is the exact FIM $F$, in the middle is our approximation $F_{\rm TKFAC}$, and on the right is the absolute error of these. The white level corresponding to the size of these absolute values.}
	\label{error-1}
\end{figure}

In this subsection, we discuss the approximation error of TKFAC and give some comparisons. Firstly, we give a visualization result of TKFAC's approximation error on MNIST in Figure \ref{error-1}. Then, we consider the approximation error $\|F-F_{\rm {TKFAC}}\|_F$ \footnote{For simplicity, we omit the subscript $l$ in this subsection.} in Theorem \ref{apperr}, where $F$ is the exact FIM defined by Eq. (\ref{fc1}), $F_{\rm {TKFAC}}$ is defined by Eq. (\ref{mod}) and $\|\cdot\|_F$ is the Frobenius norm of a matrix. We give a upper bound of TKFAC's approximation error. Finally, we compare the approximation errors of KFAC and TKFAC in experiments.

Figure \ref{error-1} shows the visualization result of the exact FIM $F$ (on the left), TKFAC's approximation $F_{\rm TKFAC}$ (in the middle) and the absolute error of these (on the right). We can see that $F_{\rm TKFAC}$ is very close to the exact $F$ and the approximation error is small. So TKFAC's approximation is efficient.

Next, we give an upper bound of TKFAC's approximation error.

\begin{theorem}\label{apperr}
Suppose that $F^{(i)}=a^{(i)}a^{(i)\top}\otimes g^{(i)}g^{(i)\top}=\Lambda^{(i)}\otimes\Gamma^{(i)}$, $i\in \{1,2,\cdots,N\}$ be the FIM of different inputs (here $N$ is the batch-size). Let $F$ be the exact FIM defined by Eq. (\ref{fc1}) and $F_{\rm {TKFAC}}$ be the approximate FIM in TKFAC defined by Eq. (\ref{mod}). We have
\[
\|F-F_{\rm {TKFAC}}\|_F\leq
\frac{2(N-1)}{N}\max\limits_{i<j \atop i,j\in\{1,2,\cdots,N\}}\left\{\sqrt{{\rm tr}(\Lambda^{(i)}){\rm tr}(\Lambda^{(j)}){\rm tr}(\Gamma^{(i)}){\rm tr}(\Gamma^{(j)})}\right\}.
\]
\end{theorem}

To prove Theorem \ref{thm1}, we need to introduce a simple lemma.

\begin{lemma}\label{equ}
Suppose that $a_1,a_2,\cdots, a_N$ and $b_1,b_2,\cdots, b_N$ are positive numbers, then the following inequality holds.
\[
\frac{a_1+a_2+ \cdots +a_N}{b_1+b_2+ \cdots +b_N}\leq \max\limits_{ i\in\{1,2,\cdots,N\}}\left\{\frac{a_i}{b_i}\right\}.
\]
\end{lemma}

Now, we give the proof of Theorem \ref{apperr}.
\begin{proof}
By the define of $F$ and $\|F-F_{\rm {TKFAC}}\|_F$, we have
\[
F=\frac{\sum_{i=1}^{N}(\Lambda^{(i)}\otimes\Gamma^{(i)})}{N},\quad
F_{\rm {TKFAC}}=\frac{\sum_{i=1}^{N}({\rm tr}(\Gamma^{(i)})\Lambda^{(i)})\otimes \sum_{i=1}^{N}({\rm tr}(\Lambda^{(i)})\Gamma^{(i)})}
{N\sum_{i=1}^{N}({\rm tr}(\Lambda^{(i)}){\rm tr}(\Gamma^{(i)}))}.
\]
Then,

\begin{align*}
\begin{aligned}
&\quad\|F-F_{\rm {TKFAC}}\|_F\\
  &=\left\|
  \frac{\sum\limits_{i=1}^{N}\Lambda^{(i)}\otimes\Gamma^{(i)}}{N}-
  \frac{\sum\limits_{i=1}^{N}{\rm tr}(\Gamma^{(i)})\Lambda^{(i)}\otimes \sum\limits_{i=1}^{N}{\rm tr}(\Lambda^{(i)})\Gamma^{(i)}}
{N\sum\limits_{i=1}^{N}{\rm tr}(\Lambda^{(i)}){\rm tr}(\Gamma^{(i)})}
  \right\|_F\\
  &=
  \frac{\left\|\sum\limits_{i=1}^{N}{\rm tr}(\Lambda^{(i)})\Lambda^{(i)}\otimes \sum\limits_{i=1}^{N}{\rm tr}(\Gamma^{(i)})\Gamma^{(i)}
 -\sum\limits_{i=1}^{N}
 {\rm tr}(\Gamma^{(i)})\Lambda^{(i)}\otimes
 \sum\limits_{i=1}^{N}{\rm tr}(\Lambda^{(i)})\Gamma^{(i)} \right\|_F}
 {N\sum\limits_{i=1}^{N}{\rm tr}(\Lambda^{(i)}){\rm tr}(\Gamma^{(i)})}\\
 &=
 \frac{\left\|\sum\limits_{i<j,\;{\rm and}\; i,j\in\{1,2,\cdots,N\}}^{N}
 \left({\rm tr}(\Lambda^{(j)})\Lambda^{(i)}-{\rm tr}(\Lambda^{(i)})\Lambda^{(j)}\right)
    \otimes
     \left({\rm tr}(\Gamma^{(i)})\Gamma^{(j)}-{\rm tr}(\Gamma^{(j)})\Gamma^{(i)}\right)\right\|_F
 }
   {N\sum\limits_{i=1}^{N}{\rm tr}(\Lambda^{(i)}){\rm tr}(\Gamma^{(i)})}\\
 &\leq
 \frac{\sum\limits_{i<j,\;{\rm and}\; i,j\in\{1,2,\cdots,N\}}^{N}
 \left\|{\rm tr}(\Lambda^{(j)})\Lambda^{(i)}-{\rm tr}(\Lambda^{(i)})\Lambda^{(j)}\right\|_F
 \left\|{\rm tr}(\Gamma^{(i)})\Gamma^{(j)}-{\rm tr}(\Gamma^{(j)})\Gamma^{(i)}\right\|_F
 }
   {N\sum\limits_{i=1}^{N}{\rm tr}(\Lambda^{(i)}){\rm tr}(\Gamma^{(i)})}\\
  &=\frac{N-1}{N}
 \frac{\sum\limits_{i<j,\;{\rm and}\; i,j\in\{1,2,\cdots,N\}}^{N}
 \left\|{\rm tr}(\Lambda^{(j)})\Lambda^{(i)}-{\rm tr}(\Lambda^{(i)})\Lambda^{(j)}\right\|_F
 \left\|{\rm tr}(\Gamma^{(i)})\Gamma^{(j)}-{\rm tr}(\Gamma^{(j)})\Gamma^{(i)}\right\|_F
 }
   {(N-1)\sum\limits_{i=1}^{N}
   {\rm tr}(\Lambda^{(i)}){\rm tr}(\Gamma^{(i)})}.\\
\end{aligned}
\end{align*}
By Lemma \ref{equ}, we have
\begin{align*}
\begin{aligned}
&\frac{N-1}{N}
 \frac{\sum\limits_{i<j,\;{\rm and}\; i,j\in\{1,2,\cdots,N\}}^{N}
 \left\|{\rm tr}(\Lambda^{(j)})\Lambda^{(i)}-{\rm tr}(\Lambda^{(i)})\Lambda^{(j)}\right\|_F
 \left\|{\rm tr}(\Gamma^{(i)})\Gamma^{(j)}-{\rm tr}(\Gamma^{(j)})\Gamma^{(i)}\right\|_F
 }
   {(N-1)\sum\limits_{i=1}^{N}
   {\rm tr}(\Lambda^{(i)}){\rm tr}(\Gamma^{(i)})}\\
 &\leq
 \frac{N-1}{N}\max\limits_{i<j \atop i,j\in\{1,2,\cdots,N\}}
 \left\{
 \frac{
 \left\|{\rm tr}(\Lambda^{(j)})\Lambda^{(i)}-{\rm tr}(\Lambda^{(i)})\Lambda^{(j)}\right\|_F
 \left\|{\rm tr}(\Gamma^{(i)})\Gamma^{(j)}-{\rm tr}(\Gamma^{(j)})\Gamma^{(i)}\right\|_F
 }
   {
   {\rm tr}(\Lambda^{(i)}){\rm tr}(\Gamma^{(i)})+(\Lambda^{(j)}){\rm tr}(\Gamma^{(j)})}
 \right\}\\
 &\leq
 \frac{N-1}{N}\max\limits_{i<j \atop i,j\in\{1,2,\cdots,N\}}
 \left\{
 \frac{{\rm tr}(\Lambda^{(j)})\|\Lambda^{(i)}\|_F+{\rm tr}(\Lambda^{(i)})\|\Lambda^{(j)}
    \|_F
 }
   {\sqrt{2{\rm tr}(\Lambda^{(i)}){\rm tr}(\Lambda^{(j)})}}
 \frac{
   {\rm tr}(\Gamma^{(j)})\|\Gamma^{(i)}\|_F+{\rm tr}(\Gamma^{(i)})\|\Gamma^{(j)}\|_F
 }
   {\sqrt{2{\rm tr}(\Gamma^{(i)}){\rm tr}(\Gamma^{(j)})}}
 \right\}.\\
 \end{aligned}
\end{align*}
According to the relationship of the Frobenius norm and the nuclear norm for any matrix $A$, which is given as
$ \|A\|_F \leq \|A\|_{\ast}, $
where $\|\cdot\|_*$ is the nuclear norm of a matrix, i.e., the sum of singular values (\citeauthor{thm2005}, \citeyear{thm2005}; \citeauthor{thm2008}, \citeyear{thm2008}). What's more, for any matrix $A$,  $|{\rm tr}(A)|$ is also smaller or equal the nuclear norm $\|A\|_*$ and the equality holds if $A$ is a positive semidefinite matrix. Therefore, $\|\Lambda^{(i)}\|_F\leq {\rm tr}(\Lambda^{(i)})$ and $\|\Gamma^{(i)}\|_F\leq {\rm tr}(\Gamma^{(i)})$ for $i=1,2,\cdots,N$. So we can obtain
\begin{align*}
\begin{aligned}
 &\frac{N-1}{N}\max\limits_{i<j \atop i,j\in\{1,2,\cdots,N\}}
 \left\{
 \frac{{\rm tr}(\Lambda^{(j)})\|\Lambda^{(i)}\|_F+{\rm tr}(\Lambda^{(i)})\|\Lambda^{(j)}
    \|_F
 }
   {\sqrt{2{\rm tr}(\Lambda^{(i)}){\rm tr}(\Lambda^{(j)})}}
 \frac{
   {\rm tr}(\Gamma^{(j)})\|\Gamma^{(i)}\|_F+{\rm tr}(\Gamma^{(i)})\|\Gamma^{(j)}\|_F
 }
   {\sqrt{2{\rm tr}(\Gamma^{(i)}){\rm tr}(\Gamma^{(j)})}}
 \right\}\\
 &\leq
  \frac{N-1}{N}\max\limits_{i<j \atop i,j\in\{1,2,\cdots,N\}}
 \left\{
 \frac{{\rm tr}(\Lambda^{(j)}){\rm tr}(\Lambda^{(i)})+{\rm tr}(\Lambda^{(i)}){\rm tr}(\Lambda^{(j)})
 }
   {\sqrt{2{\rm tr}(\Lambda^{(i)}){\rm tr}(\Lambda^{(j)})}}
 \frac{
   {\rm tr}(\Gamma^{(j)}){\rm tr}(\Gamma^{(i)})+{\rm tr}(\Gamma^{(i)}){\rm tr}(\Gamma^{(j)})
 }
   {\sqrt{2{\rm tr}(\Gamma^{(i)}){\rm tr}(\Gamma^{(j)})}}
 \right\}\\
 &\leq
 \frac{2(N-1)}{N}
 \max\limits_{i<j \atop i,j\in\{1,2,\cdots,N\}}\left\{\sqrt{{\rm tr}(\Lambda^{(i)}){\rm tr}(\Lambda^{(j)}){\rm tr}(\Gamma^{(i)}){\rm tr}(\Gamma^{(j)})}\right\}.
\end{aligned}
\end{align*}

The proof is complete.
\end{proof}

We also consider the KFAC approximation to the FIM, which is give by
$$F_{\rm KFAC}=\mathbb{E}[\Lambda]\otimes\mathbb{E}[\Gamma]=\frac{\sum_{i=1}^N \Lambda^{(i)}}{N}\otimes\frac{\sum_{i=1}^N \Gamma^{(i)}}{N}.
$$
Similar to the discuss of $\|F-F_{\rm {KFAC}}\|_F$, we can also obtain an upper bound of the the approximation error $\|F-F_{\rm {KFAC}}\|_F$ under the same proof process as Theorem \ref{apperr}.
$$\|F-F_{\rm {KFAC}}\|_F\leq
\frac{2(N-1)}{N}\max\limits_{i<j \atop i,j\in{1,2,\cdots,N}}
\left\{\frac{({\rm {tr}}(\Lambda^{(i)})+{\rm {tr}}(\Lambda^{(j)}))({\rm {tr}}(\Gamma^{(i)})+{\rm {tr}}(\Gamma^{(j)}))
 }{4}
 \right\}
$$

\begin{figure}[htb]
  \vspace{-0.35cm}
  \centering
  \includegraphics[width=0.35\linewidth]{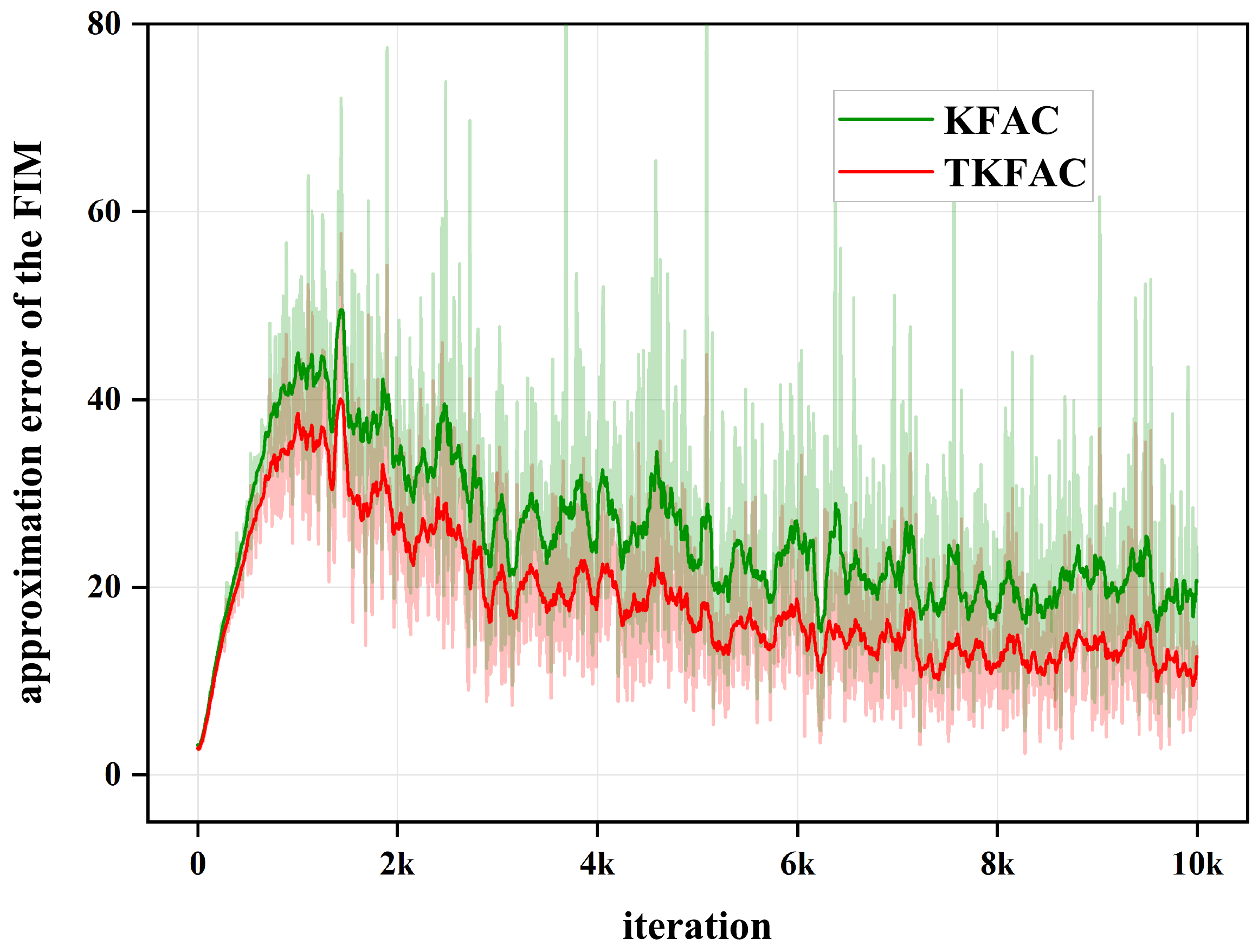}\\
  \caption{The curve of the sum of each layer's approximation error with iteration for KFAC and TKFAC. The model is same as described in Figure \ref{error-1}. }\label{error-2}
\end{figure}

Note that in general cases except that ${\rm {tr}}(\Lambda^{(i)}={\rm {tr}}(\Lambda^{(j)})$ for all $i\in\{1,2,\cdots,N\}$, the upper bound of TKFAC is smaller than KFAC. But this does not mean that TKFAC's approximation error is smaller than KFAC. Such a result may be difficult to obtain in theory, so we compare the approximation errors of these two methods in experiment.

In Figure \ref{error-2}, we show the comparison of $\|F-F_{\rm {KFAC}}\|_F$ and $\|F-F_{\rm {TKFAC}}\|_F$ when training MNIST, and we give the curve of the sum of each layer's error with iterations. It can be seen that $\|F-F_{\rm {TKFAC}}\|_F$ is smaller than $\|F-F_{\rm {KFAC}}\|_F$, which indicates that our approximation may be more exact than KFAC in this case.

\section{Using TKFAC for training}\label{sec-5}

To use TKFAC for training DNNs, some tricks should be employed. In this section, we mainly introduce the normal damping technique, a new damping technique for TKFAC on CNNs and the exponential moving averages. We also give the pseudocode of TKFAC in Algorithm \ref{alg1}.

\noindent{\bf The normal damping technique:} \citet{thm2019} showed that most eigenvalues of the FIM of DNNs are close to zero, while only a small number of eigenvalues take on large values. This leads to most eigenvalues of the inverse FIM to be extremely huge or even infinite, which causes computational difficulty and inaccuracy. To make TKFAC stable, we add $\lambda I$ to the FIM, and we will take the fully-connected layer as an example to illustrate our damping method in the following. For the fully-connected layer, we add $\lambda I_{m_{l-1}m_l}\in \mathbb{R}^{m_{l-1}m_{l}\times m_{l-1}m_{l}}$ into $F_{l}=\in \mathbb{R}^{m_{l-1}m_l\times m_{l-1}m_l}$, i.e.,
\begin{equation}\label{dam1}
  F_{l}+\lambda I_{m_{l-1}m_l}=\delta_l\Phi_l\otimes\Psi_l+\lambda I_{m_{l-1}}\otimes I_{m_l}.
\end{equation}
In order to invert $F_l+\lambda I_{m_{l-1}m_l}$ easily in computation using the properties of the Kronecker product, we approximate it by the following formula, which has been proposed in \citet{kfac2015}.
\begin{equation}\label{dam2}
  \hat{\Phi}_l=\sqrt{\delta_l}\Phi_l+\sqrt{\lambda}I_{m_{l-1}}, \quad
  \hat{\Psi}_l=\sqrt{\delta_l}\Psi_l+\sqrt{\lambda}I_{m_{l}}.
\end{equation}
where $\delta_l,\Phi_l$ and $\Psi_l$ are defined by Eq. (\ref{sim}) (Eq. (\ref{th2-3}) for convolutional layers).

\begin{figure}[htb]
	\centering
	\vspace{-0.35cm}
	\subfigtopskip=2pt
	\subfigbottomskip=2pt
	\subfigcapskip=2pt
	\subfigure[
       ]{
		\label{damp-1}
		\includegraphics[width=0.3\linewidth]{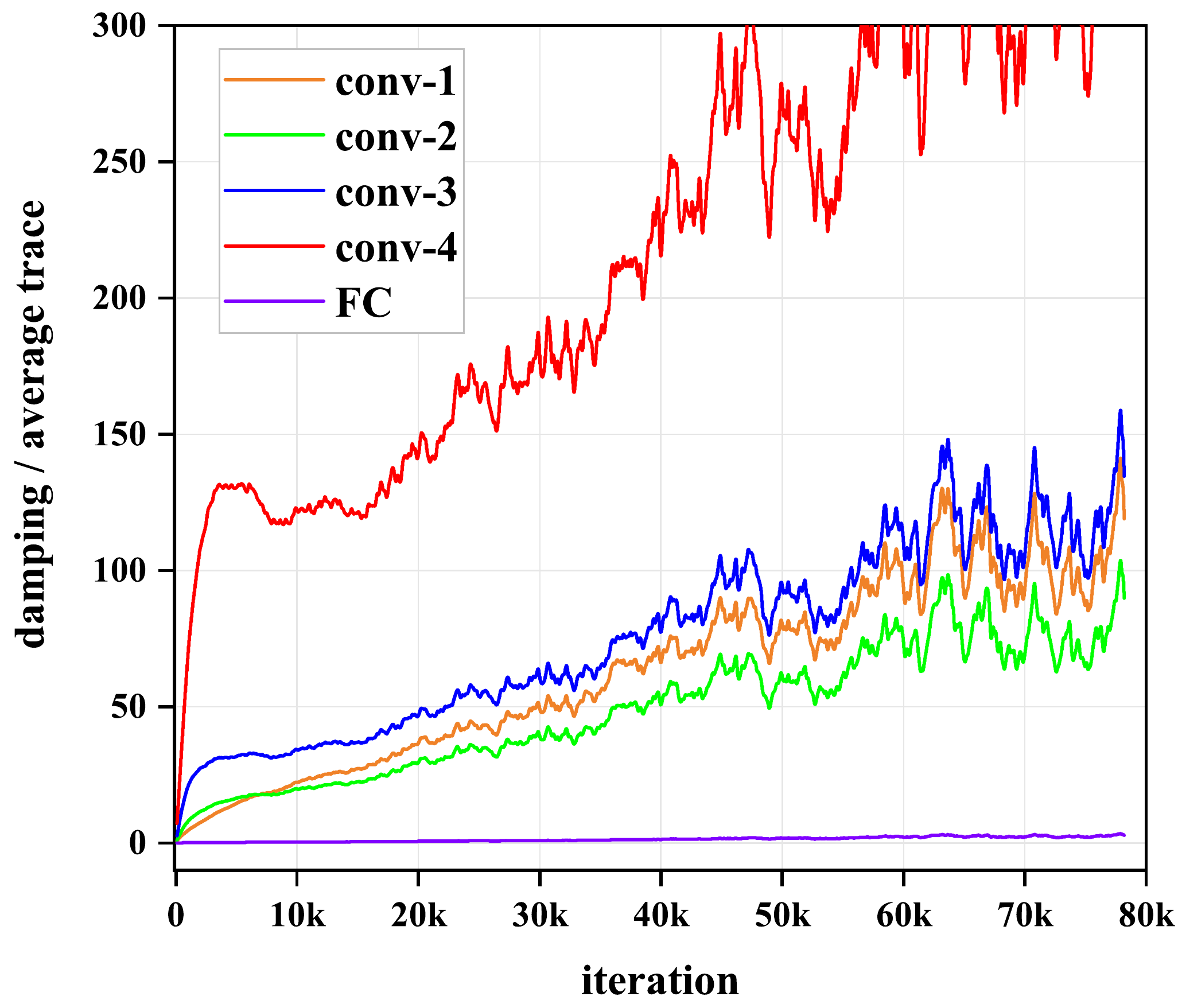}}
	\subfigure[
        ]{
		\label{damp-2}
		\includegraphics[width=0.3\linewidth]{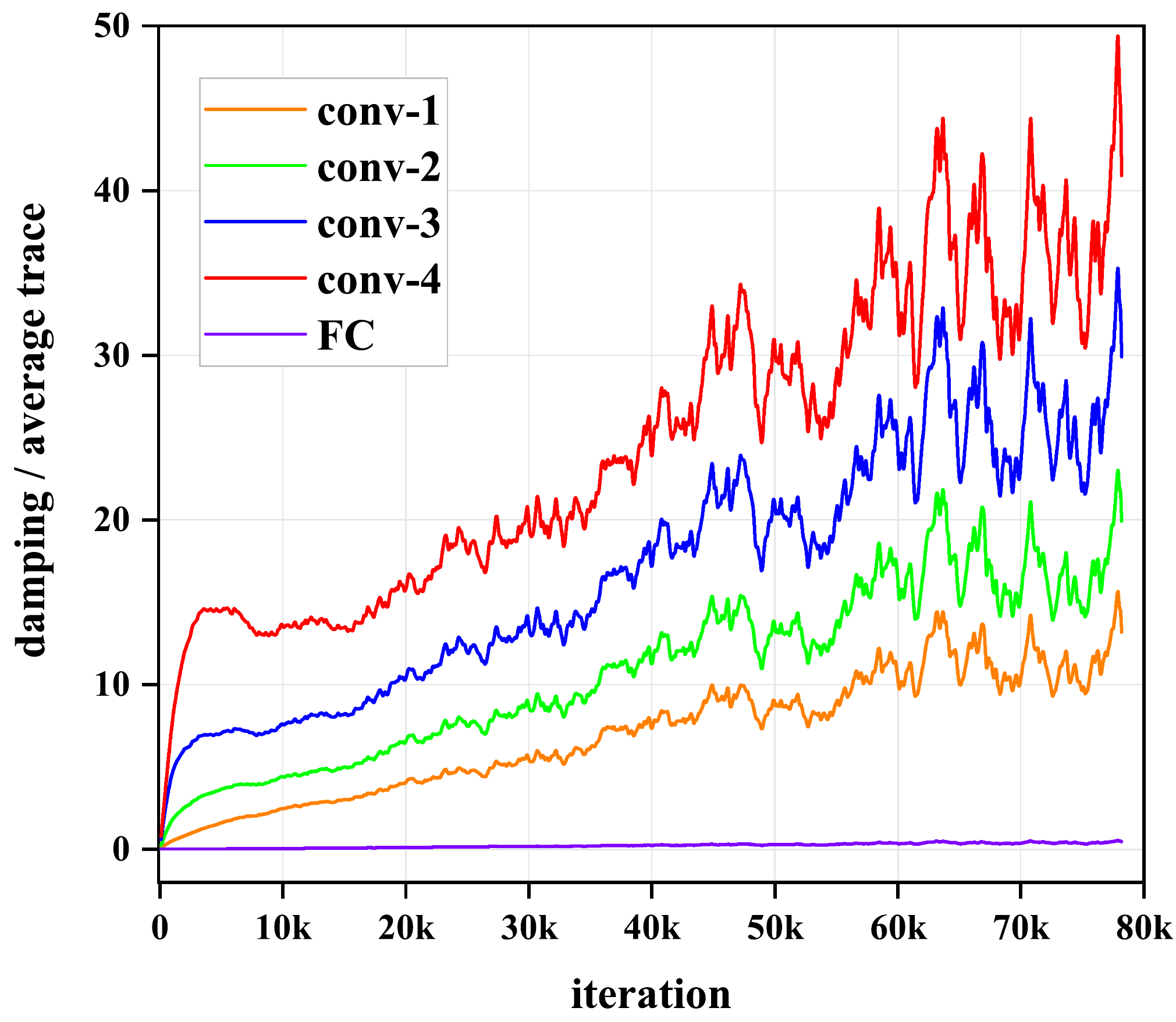}}
    \subfigure[
        ]{
		\label{damp-3}
		\includegraphics[width=0.3\linewidth]{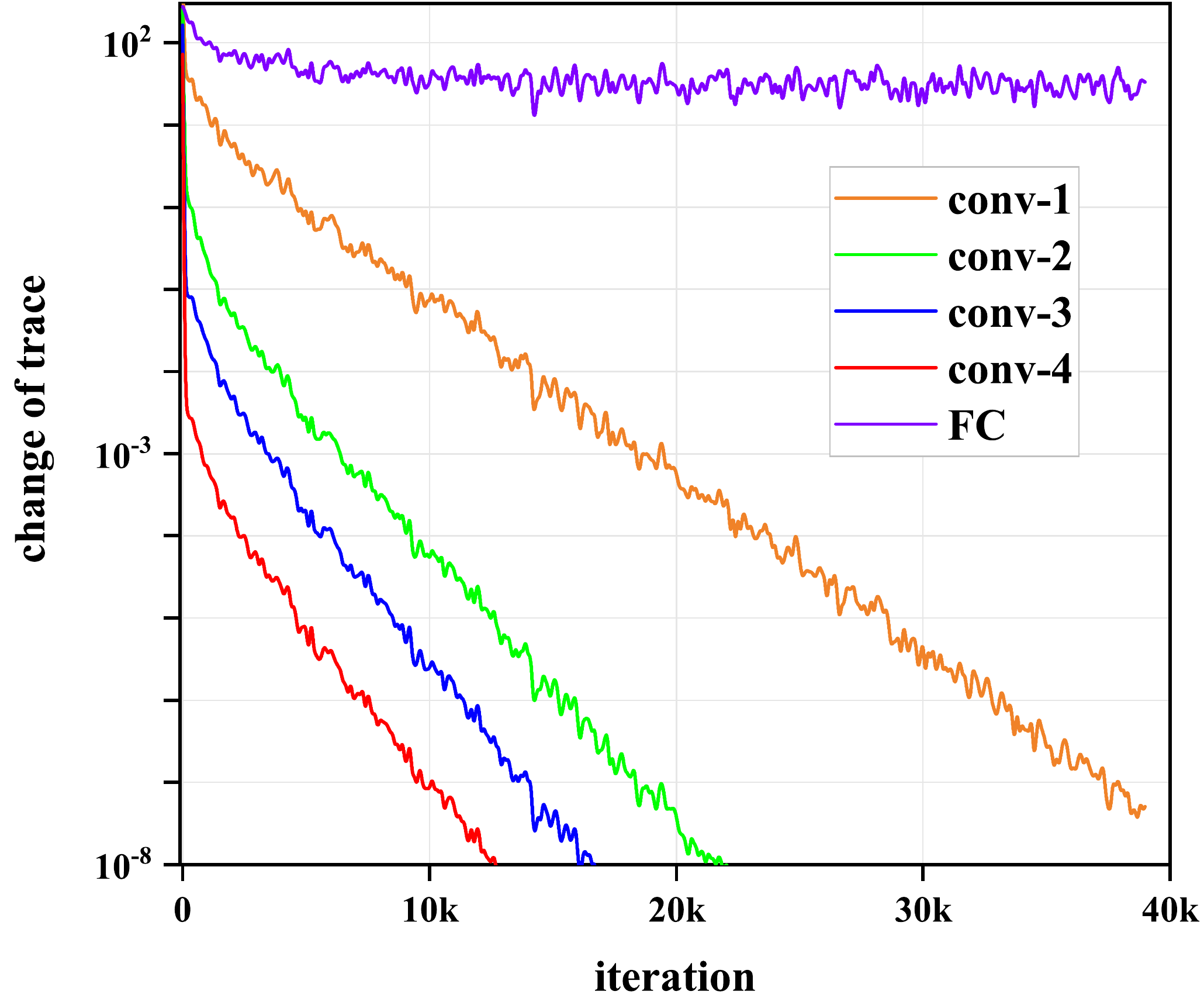}}
\caption{Some comparison results. We choose four different convolutional layers and the fully-connected layer from ResNet20 network when training CIFAR-10 using TKFAC. We record data every 100 iterations. In (a), we show the changes of $\sqrt{\lambda}$/(tr($\delta_l\Phi_l$)/$m_{l-1}$). In (b), we show the changes of $\sqrt{\lambda}$/(tr($\delta_l\Psi_l$)/$m_{l}$). In (c), we show the changes of traces without restricting $\delta_l$.}
	\label{damp}
\end{figure}

\noindent{\bf A new damping technique for CNNs:}
In experiments, we find that this damping technique has some limitations for convolutional layers. In Figure \ref{damp}, we compare the two ratios, which are the damping divided by the mean of the diagonal elements of $\delta_l\Phi_l$ and $\delta_l\Psi_l$ for some layers of ResNet20 when training CIFAR-10 using TKFAC. As shown in Figure \ref{damp-1} and Figure \ref{damp-2}, we find that the damping of the convolutional layers will soon be much larger than the mean of its diagonal elements, which means the damping may play a major role rather than the FIM shortly after the start of training. But this problem doesn't exist in the fully-connected layer as shown by the purple curves. This may limit to use the information of the FIM adequately. So we adopt the following damping technique for convolutional layers, which has good performance in our experiments. That is
\begin{equation}\label{dam3}
  \tilde{\delta}_l=\max\{\nu, \delta_l\},\quad
  \tilde{\Phi}_l=\sqrt{\tilde{\delta}_l}\Phi_l+\frac{\tilde{\delta}_l}{n_{l-1}k_l^2}I_{n_{l-1}k_l^2\times n_{l-1}k_l^2},
  \quad \tilde{\Psi}_l=\sqrt{\tilde{\delta}_l}\Psi_l+\frac{\tilde{\delta}_l}{n_l}I_{n_l\times n_l}.
\end{equation}
where $\delta_l,\Phi_l$ and $\Psi_l$ are defined by Eq. (\ref{th3}) and $\nu$ is a reasonably large positive scalar.

The reason why we restrict $\delta_l$ is that $\delta_l$ may become very small and can't promise that $\Phi_l$ and $\Psi_l$ don't have very small eigenvalues. In Figure \ref{damp-3}, we choose the traces of the fully-connected layer and four different convolutional layers from ResNet20 when training CIFAR-10 using TKFAC with this damping technique but without restricting $\delta_l$. We can see that the values of $\delta_l$ are very different, and some values are very small or even become close to zero. So the training becomes unstable and a constraint on $\delta_l$ is necessary. What's more, we also notice that in Figure \ref{damp-3}, $\delta_l$ of convolutional layers and the fully-connected layer vary widely. In fact, the role of $\nu$ is to expand the elements of the FIM for convolutional layers by some times. However, this expansion does not work on the fully-connected layer, because the trace of the fully-connected layer is greater than $\nu$. In order to keep pace with convolutional layers, we also expanded the FIM of the fully-connected layer by a factor of $\beta$, where
\begin{equation}\label{tr2}
  \beta=\max\limits_{l\in\{1, \cdots, L\}}[\tilde{\delta}_l/\delta_l].
\end{equation}

\noindent{\bf Exponential moving averages:} We also use an exponential moving averages as previous works, which take the new estimate to be the old one weighted by $\varepsilon$, plus the estimate computed on the new mini-batch weighted by $1-\varepsilon$. That is
\begin{equation}\label{av1}
  \hat{\Phi}_l^{(t+1)}\leftarrow\varepsilon\hat{\Phi}_l^{(t)}+(1-\varepsilon)\hat{\Phi}_l^{(t+1)},
\end{equation}
\begin{equation}\label{av2}
  \hat{\Psi}_l^{(t+1)}\leftarrow\varepsilon\hat{\Psi}_l^{(t)}+(1-\varepsilon)\hat{\Psi}_l^{(t+1)},
\end{equation}
where $\hat{\Phi}_l$ and $\hat{\Psi}_l$ are computed by Eq. (\ref{dam2}) (It's similar to $\tilde{\Phi}_l$ and $\tilde{\Psi}_l$ if we use the new damping technique) and $t$ is the iteration.
Finally, the parameters are updated
$$\omega_l^{(t+1)}\leftarrow\omega_l^{(t)}-\alpha ((\hat{\Phi}_l^{(t)})^{-1}\otimes (\hat{\Psi}_l^{(t)})^{-1})\nabla_{\omega_l}h^{(t)}.$$

Algorithm \ref{alg1} gives a high level pseudocode of TKFAC with normal damping technique. TKFAC with the new damping technique (only for CNNs) is similar to algorithm \ref{alg1} except the differences as follows: a) We adopt the damping technique defined by Eq. (\ref{dam3}) rather than Eq. (\ref{dam2}), and the damping does not need to be tuned as a hyperparameter during training. b) We use a parameter $\nu$ to restrict traces given in Eq. (\ref{dam3}), and expand the FIM of the fully-connected layer by a factor of $\beta$ defined by Eq. (\ref{tr2}).

\begin{algorithm}[H]
\caption{TKFAC}\label{alg1}
\begin{algorithmic}
\REQUIRE $\alpha$ : learning rate
\REQUIRE $\lambda$ : damping parameter
\REQUIRE $\tau$ : momentum parameter
\REQUIRE $\varepsilon$ : exponential moving average parameter
\REQUIRE ${\rm T_{FIM}, T_{INV}}$ : FIM and its inverse update intervals
\STATE   $t\leftarrow0$, $m\leftarrow0$, Initialize $\{\hat{\Phi}_l\}_{l=1}^L$ and $\{\hat{\Psi}_l\}_{l=1}^L$
\WHILE{convergence is not reached}
\STATE Select a new mini-batch
\FORALL{$l\in \{1,2,\cdots,L\}$ }
\IF{$t\equiv0$ (mod ${\rm T_{FIM}}$)}
\STATE Update the factors $\hat{\Phi}_l$ and $\hat{\Psi}_l$ 
\ENDIF
\IF{$t\equiv0$ (mod ${\rm T_{INV}}$)}
\STATE Compute the inverses of $\hat{\Phi}_l$ and $\hat{\Psi}_l$
\ENDIF

\STATE Compute $\nabla_{\omega_l}h$ using backpropagation
\STATE Compute the approximated natural gradient $(\hat{\Phi}_l^{-1}\otimes \hat{\Psi}_l^{-1})\nabla_{\omega_l}h$
\STATE $\zeta\leftarrow-\alpha (\hat{\Phi}_l^{-1}\otimes \hat{\Psi}_l^{-1})\nabla_{\omega_l}h$
\STATE $m\leftarrow\tau m+\zeta$ (Update momentum)
\STATE $\omega_l \leftarrow\omega_l+m$(Update parameters)
\STATE $t\leftarrow t+1$
\ENDFOR
\ENDWHILE

\RETURN $\omega$
\end{algorithmic}
\end{algorithm}

\section{Experiments} \label{sec-6}

In this section, to show the effectiveness of TKFAC, we empirically demonstrate its performance on the auto-encoder and image classification tasks. Our experiments mainly consist of two parts. In the first part, we focus on the optimization performance of TKFAC compared with other optimization methods, and in the second part, we pay more attention to its generalized performance. In these experiments, we consider two variants of TKFAC: TKFAC\_nor and TKFAC\_new, in which the normal damping technique and the new damping technique are used, respectively.

Throughout this paper, we use three different datasets, MNIST \citep{mnist1998}, CIFAR-10 and CIFAR-100 \citep{data2009}. We adopt a standard data augmentation scheme including random crop and horizontal flip for CIFAR-10 and CIFAR-100. For MNIST, we use a 8-layer FNN. For CIFAR-10 and CIFAR-100, we use VGG16 \citep{net2014} and ResNet20 \citep{net2016}. We choose SGDM, Adam \citep{adam2014} and KFAC \citep{kfac2015} as baselines. Other related methods are also considered, but they tend to have similar or worse performance than these baselines.

All experiments are run on a single RTX 2080Ti GPU using TensorFlow. We mainly follow the experimental setup without weight decay in \citet{wd2019}. The hyperparameters including the initial learning rate $\alpha$, the damping parameter $\lambda$ and the parameter $\nu$ are tuned using a grid search with values $\alpha\in$ \{1e-4, 3e-4, $\ldots$ , 1, 3\}, $\lambda\in$ \{1e-8, 1e-6, 1e-4, 3e-4, 1e-3, $\ldots$ , 1e-1, 3e-1\} and $\nu\in$ \{1e-4, 1e-3, $\ldots$ , 10\}. The moving average parameter $\varepsilon$ and the momentum are set to 0.95 and 0.9, respectively. The update intervals are set to $T_{\rm{FIM}}=T_{\rm{INV}}=100$. All experiments are run 200
epochs and repeated five times with a batch size of 500 for MNIST and 128 for CIFAR-10/100. The results are given as mean $\pm$ standard deviation.

\noindent{\textbf{Part \uppercase\expandafter{\romannumeral1}:}} In this part, we mainly focus on the optimization performance of our method, and we don't use any additional techniques. Therefore, we mainly give the results of the training set (the results of testing set are given in Appendix \ref{sec-b}) and the learning rate is kept constant during training as \citet{ekfac2018}. The TKFAC\_new is also not considered in this part.

\begin{figure}[H]
	\centering
	\vspace{-0.35cm}
	\subfigtopskip=2pt
	\subfigbottomskip=2pt
	\subfigcapskip=2pt
	\subfigure[Reconstruction error]{
		\label{mnist-1}
		\includegraphics[width=0.3\linewidth]{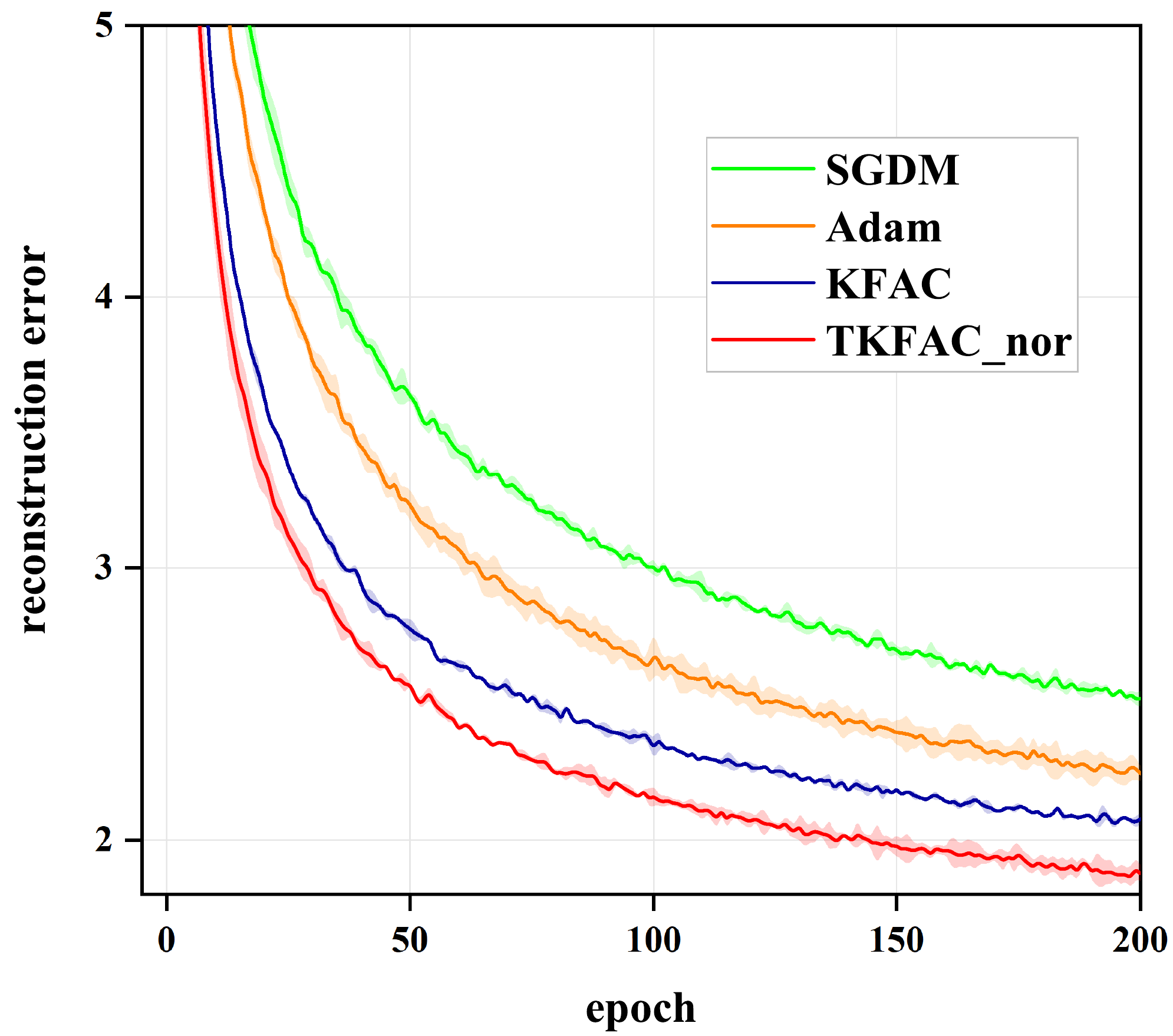}}
	\subfigure[Wall-clock time]{
		\label{mnist-2}
		\includegraphics[width=0.295\linewidth]{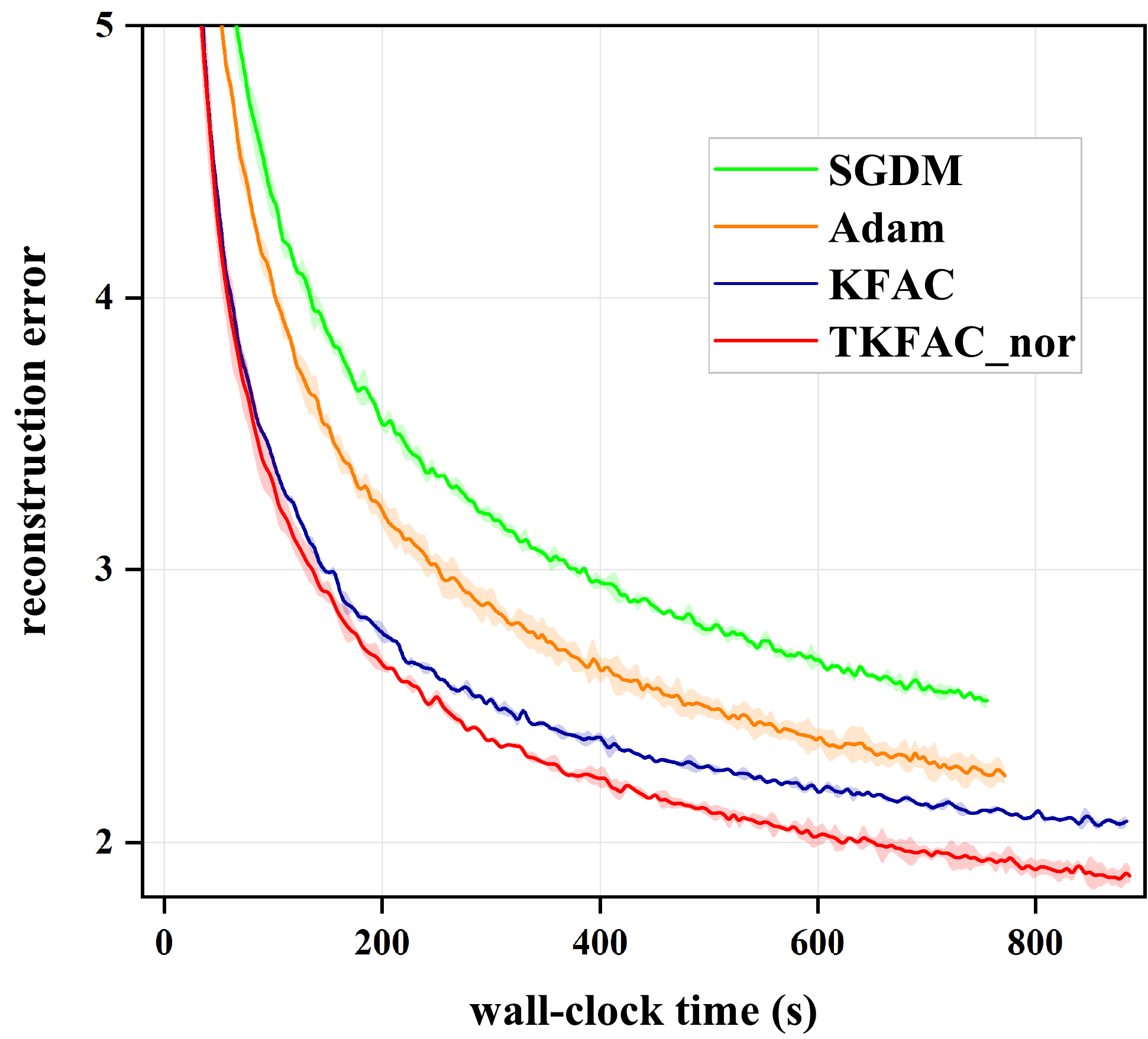}}
\caption{The curves of reconstruction error on MNIST. (a) Reconstruction error vs epochs; (b) Reconstruction error vs wall-clock time.}
	\label{mnist}
\end{figure}

We first consider a 8-layer auto-encoder task on MNIST, which is a standard task used to benchmark optimization methods (\citeauthor{kfac2015}, \citeyear{kfac2015}; \citeauthor{kfacr2018}, \citeyear{kfacr2018}; \citeauthor{qn2020}, \citeyear{qn2020}). Following these previous works, we report the reconstruction error on the training set and the results are shown in Figure \ref{mnist}. Figure \ref{mnist-1} shows the curve of error with epochs throughout training. We can see that TKFAC\_nor minimizes the error faster per epoch than other baselines and achieves lowest error after 200 epochs. As shown in Figure \ref{mnist-2}, TKFAC\_nor has similar computation time (slightly more) to KFAC and it still minimizes the error faster than other baselines in terms of time.

\begin{figure}[htb]
	\centering
	\vspace{-0.35cm}
	\subfigtopskip=2pt
	\subfigbottomskip=2pt
	\subfigcapskip=2pt
	\subfigure[Training loss]{
		\label{v16-1}
		\includegraphics[width=0.3\linewidth]{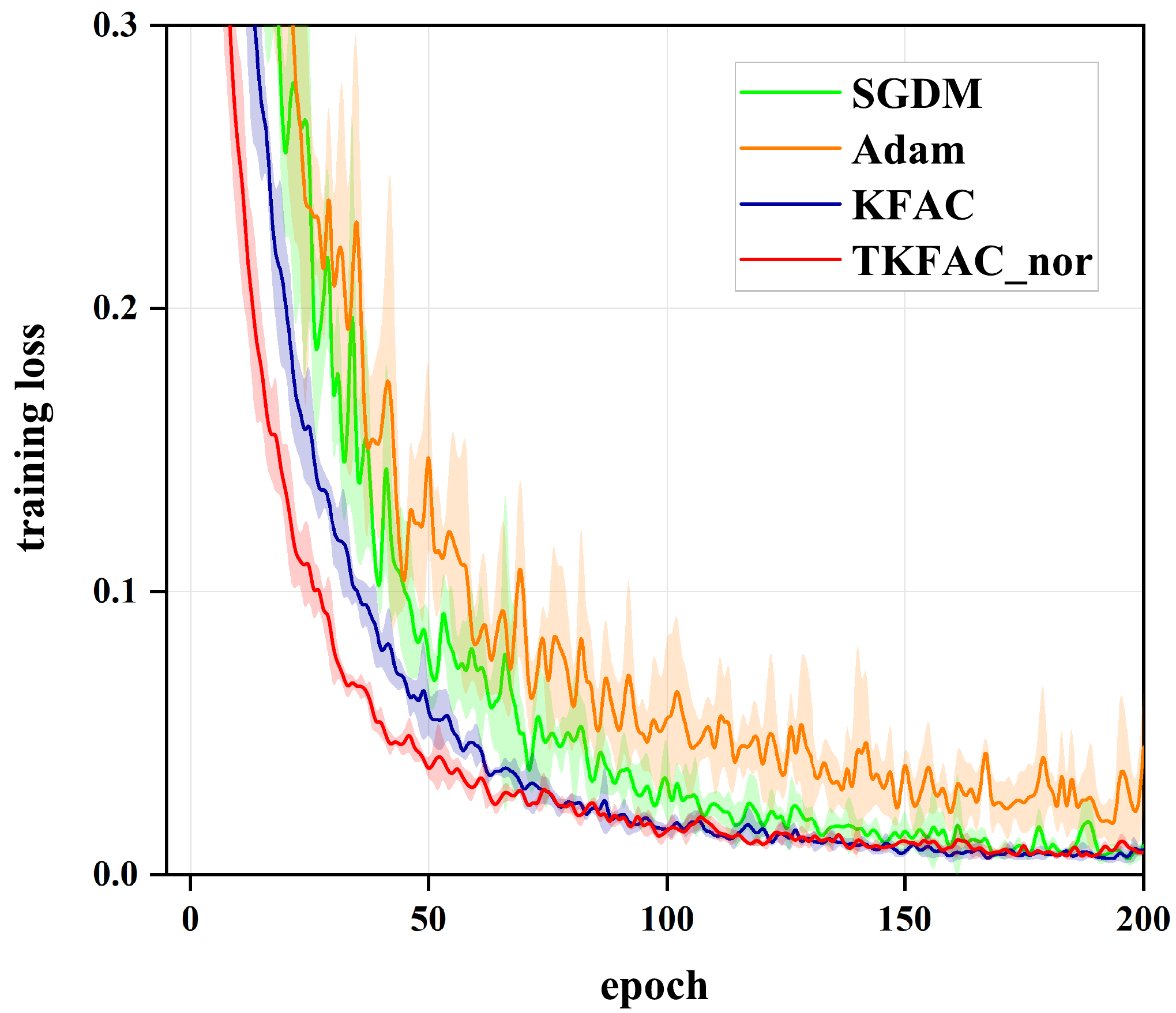}}
	\subfigure[Wall-clock time]{
		\label{v16-2}
		\includegraphics[width=0.3\linewidth]{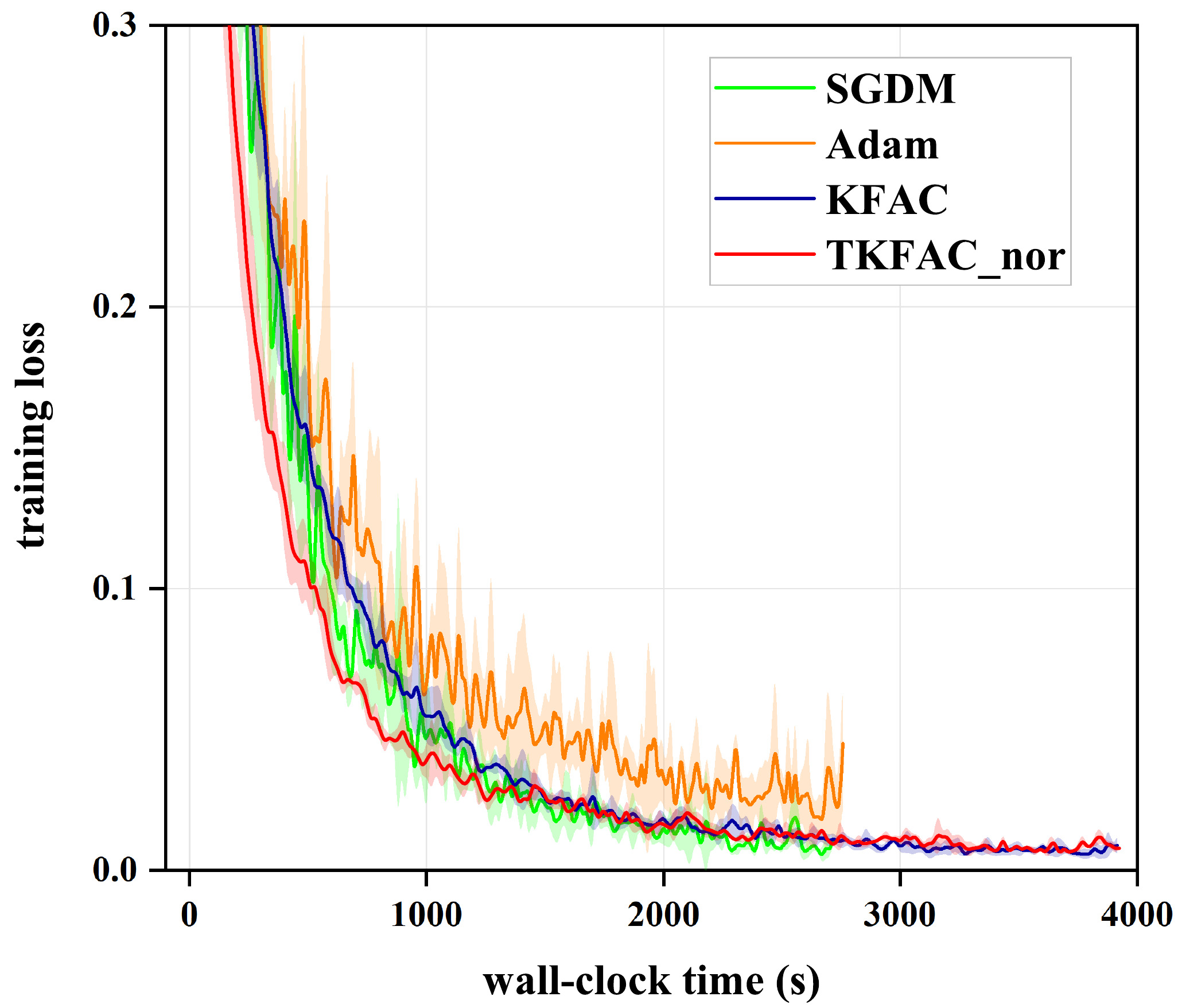}}
\caption{The curves of training loss on VGG16 for CIFAR-10. (a) Training loss vs epochs; (b) Training loss vs wall-clock time.}
	\label{v16}
\end{figure}

\begin{figure}[H]
	\centering
	\vspace{-0.35cm}
	\subfigtopskip=2pt
	\subfigbottomskip=2pt
	\subfigcapskip=2pt
	\subfigure[Testing accuracy on VGG16]{
		\includegraphics[width=0.35\linewidth]{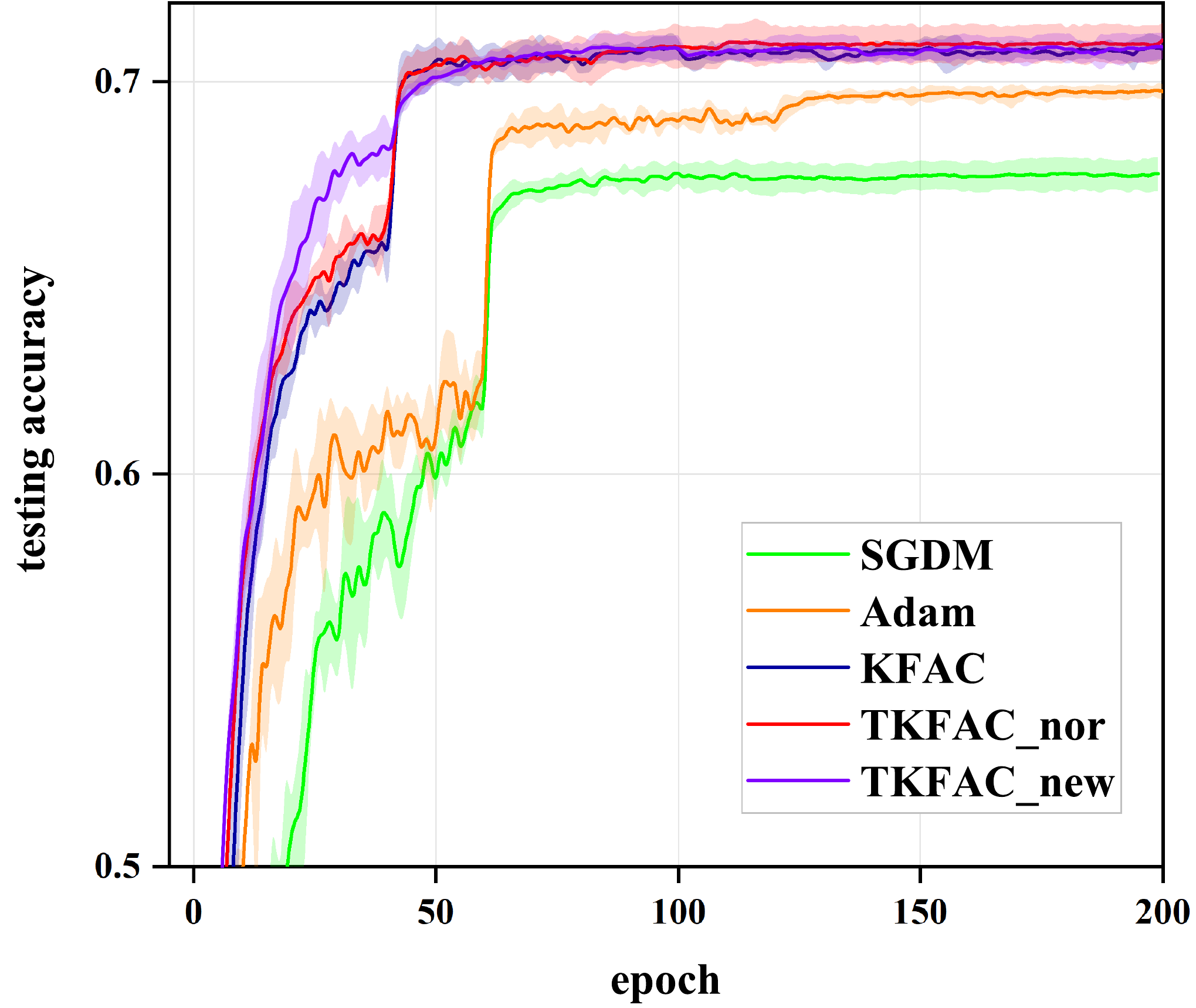}}
	\subfigure[Wall-clock time on VGG16]{
		\includegraphics[width=0.35\linewidth]{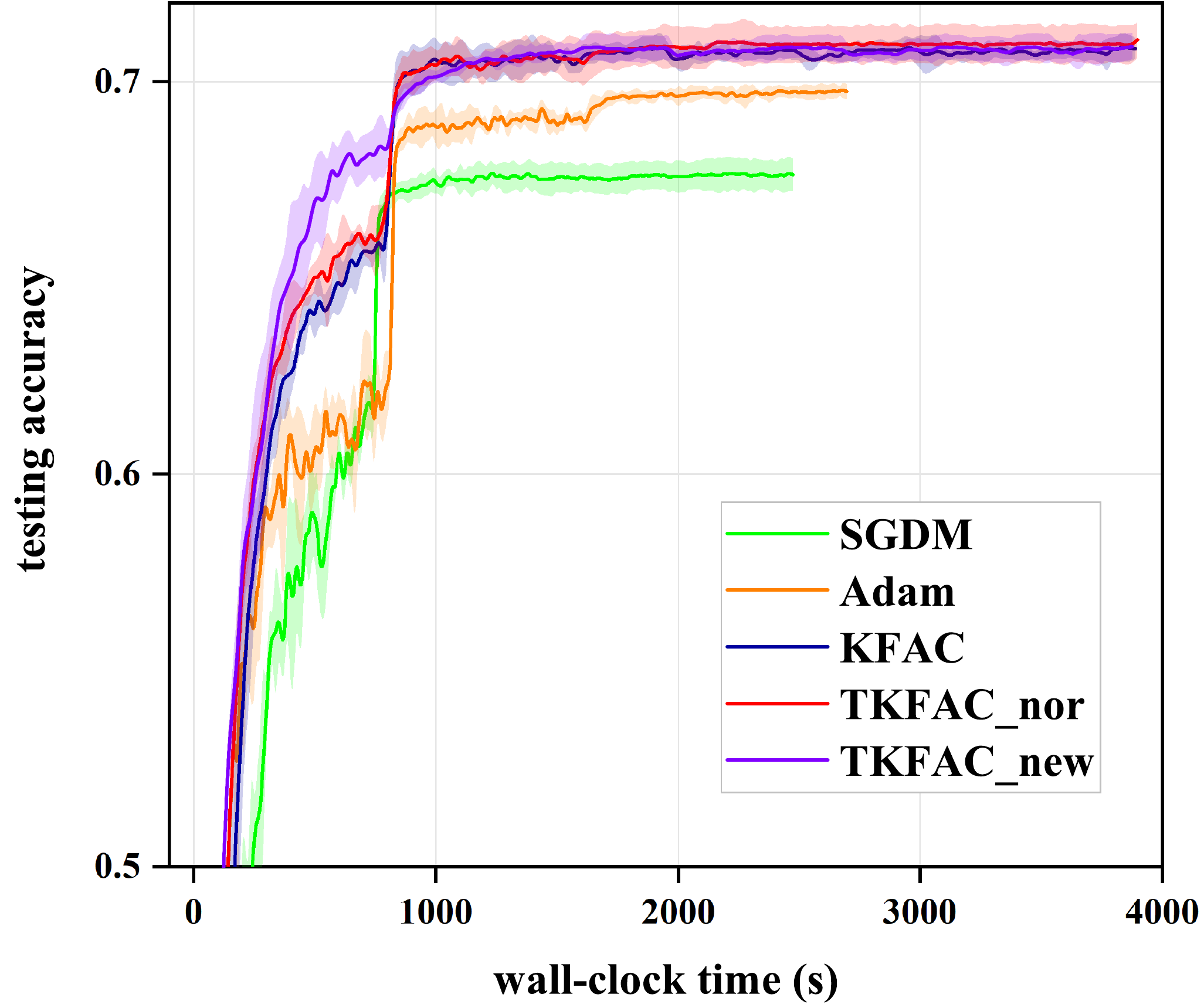}}\\
    \subfigure[Testing accuracy on ResNet20]{
		\includegraphics[width=0.35\linewidth]{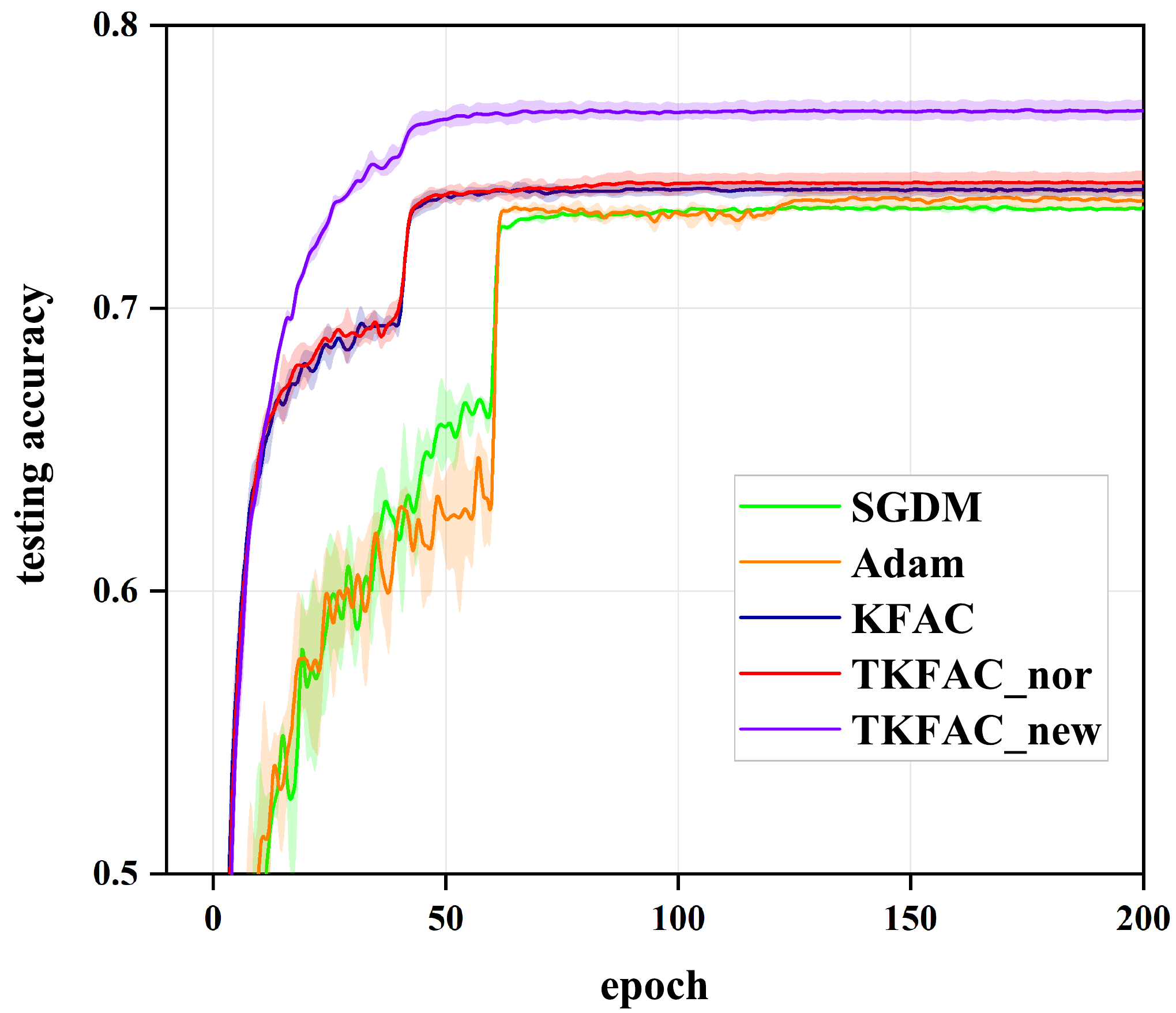}}
    \subfigure[Wall-clock time on ResNet20]{
		\includegraphics[width=0.35\linewidth]{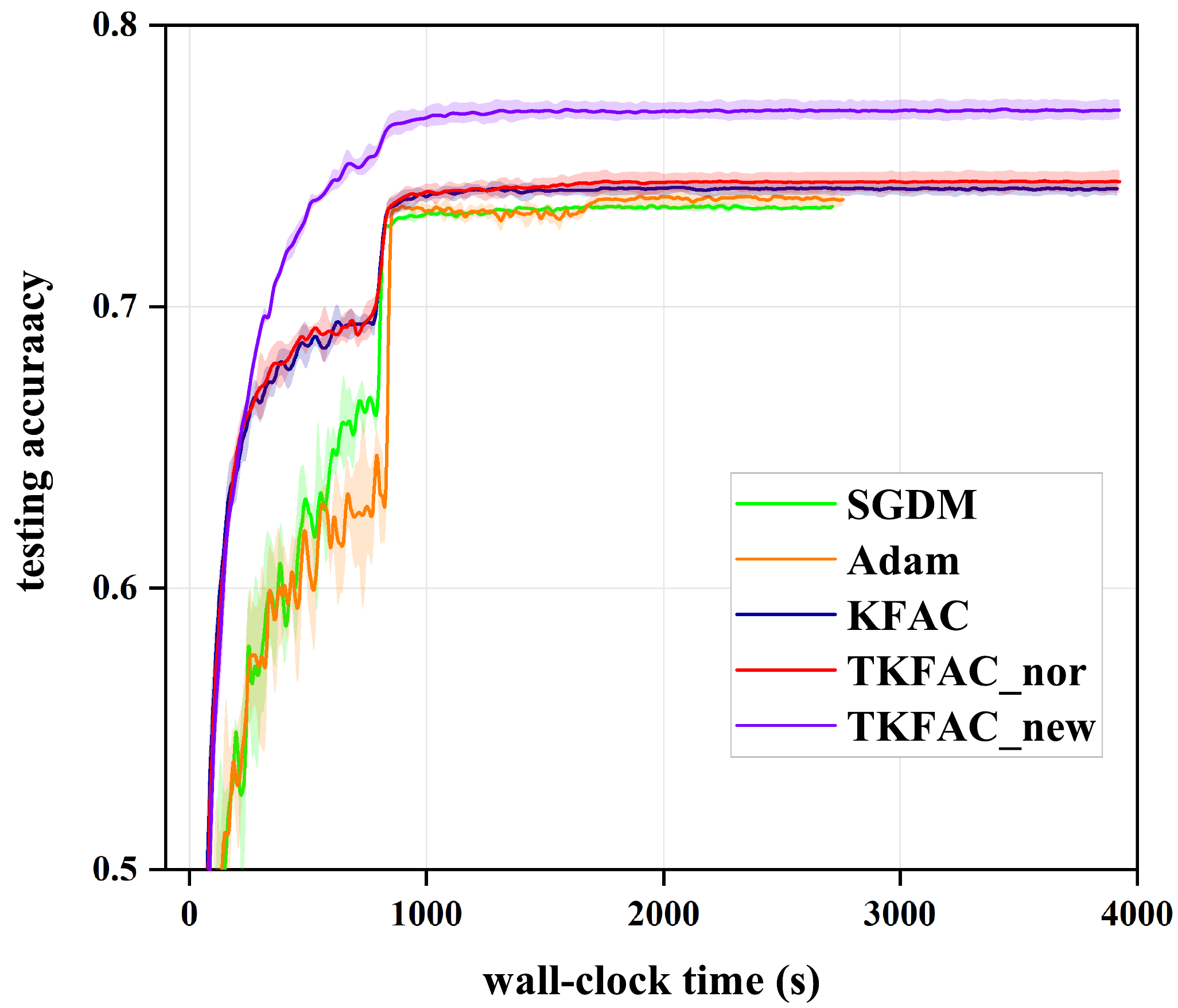}}
\caption{The curves of testing accuracy for CIFAR-100 on VGG16 and ResNet20. (a) Testing accuracy vs Epoch on VGG16; (b) Testing accuracy vs Wall-clock time on VGG16; (c) Testing accuracy vs Epoch on ResNet20; (d) Testing accuracy vs Wall-clock time on ResNet20.}
	\label{res}
\end{figure}

Next, we evaluate TKFAC on CNNs. The dataset and model used here are CIFAR-10 and VGG16, respectively. Figure \ref{v16-1} shows that TKFAC\_nor's training loss has a faster decline rate in the first few epochs. And it performs similarly to KFAC after about 70 epochs. All methods reach similar loss after 200 epochs except Adam. In terms of time, TKFAC\_nor still has some advantages over SGDM as given in \ref{v16-2}.

\noindent{\textbf{Part \uppercase\expandafter{\romannumeral2}:}} In this part, we pay more attention to evaluating TKFAC's generalized performance. We compare TKFAC\_nor, TKFAC\_new with SGDM, Adam, and KFAC on CIFAR-10 and CIFAR-100. The CNNs used are VGG16 and ResNet20. A learning rate schedule is also used. The initial learning rate is multiplied by 0.1 every 40 epochs for KFAC/TKFAC\_nor/TKFAC\_new and every 60 epochs for SGDM/Adam. Experimental results are given as follows.

Figure \ref{res} shows the curves of the testing accuracies of CIFAR100 on VGG16 and ResNet20 in terms of epoch and time. We can see that all second order optimizers (KFAC, TKFAC\_nor and TKFAC\_new) converge faster than SGDM and Adam. The convergence rate of TKFAC\_nor is slightly faster than KFAC on VGG16, but there is little difference in ResNet20. It is obvious that TKFAC\_new has a faster convergence rate than KFAC, especially on ResNet20. The final testing accuracies of these methods are given in Table \ref{acc}. TKFAC\_nor can achieve higher average accuracy than SGDM, Adam and KFAC in all cases. Especially, TKFAC\_new greatly improves the final accuracy on ResNet20 (for example, improve 2.83\% than KFAC) with Significantly faster convergence rate.

\begin{table}[htb]
\setlength{\abovecaptionskip}{0.0cm}
\setlength{\belowcaptionskip}{0.3cm}
\centering
\caption{Final testing accuracy of CIFAR-10 and CIFAR-100.}
\label{acc}
\begin{tabular}{@{}c|c|c|c@{}}
\toprule
\multicolumn{2}{c|}{Dataset}        & CIFAR-10 (\%)                & CIFAR-100 (\%)                \\ \midrule
\multirow{5}{*}{VGG16}    & SGDM    & 91.77$\pm$ 0.25          & 67.63$\pm$ 0.42          \\
                          & Adam    & 92.57$\pm$ 0.31          & 69.74$\pm$ 0.19          \\
                          & KFAC   & 92.70$\pm$ 0.39          & 70.84$\pm$ 0.27          \\
                          & TKFAC\_nor  & 93.33$\pm$ 0.20          & \textbf{71.06$\pm$0.45}       \\
                          & TKFAC\_new & \textbf{93.42$\pm$0.11} &  70.91$\pm$ 0.34    \\ \midrule
\multirow{5}{*}{ResNet20} & SGDM    & 92.78$\pm$ 0.39          & 73.55$\pm$ 0.08          \\
                          & Adam    & 93.57$\pm$ 0.22          & 73.79$\pm$ 0.21          \\
                          & KFAC   & 93.69$\pm$ 0.16          & 74.18$\pm$ 0.27          \\
                          & TKFAC\_nor  & 93.76$\pm$ 0.19          & 74.43$\pm$ 0.41         \\
                          & TKFAC\_new & \textbf{94.66$\pm$ 0.12} & \textbf{77.01$\pm$ 0.34} \\ \bottomrule
\end{tabular}
\end{table}

Experimental results show that TKFAC\_nor performs better than other baselines in most cases. TKFAC\_new can accelerate the convergence and improve the testing accuracy on ResNet20, which shows the effectiveness of our new damping technique, although it does not improve much on VGG16. Of course, due to the limitation of computational resource, the performance of TKFAC in large-scale problems needs to be further verified. At the same time, we also want to emphasize that the new damping scheme is our preliminary attempt and it may be limited by the network structure. For other second-order optimization methods, we think that their damping will have the same problem as TKFAC, that is the damping will be
large enough to dominate the curvature and turns them into first-order optimizers. So appropriate damping schemes or other techniques should be used to avoid this problem, and our approach may provide a direction.

\section{Conclusion and future work} \label{sec-7}

In this work, we presented a new method called TKFAC, where a new approximation to the FIM based on the block diagonal approximation as KFAC and a quadratic form estimator. The important property of our method is to keep the traces equal. We considered the approximation error and give some analyses. We also discussed the damping technique, and adopted a new damping technique for TKFAC on CNNs. In experiments, we showed that TKFAC have better performance than SGDM, Adam and KFAC in most cases.

Of course, we think TKFAC can also be extended to other neural networks. What's more, combined with some other tricks, TKFAC may have better performance in training. We also want to explore TKFAC's performance of more complicated training tasks. These problems are left for future work.

\bibliographystyle{plainnat}
\bibliography{d}

\appendix
\section{Implementation details}\label{sec-a}

We choose the FNN used in \citet{qn2020} for the auto-encoder task. The network architecture is 1000-500-250-30-30-250-500-1000 with the ReLU activation. The loss function is binary entropy.

For KFAC, we mainly refer to the setting in \citet{wd2019}\footnote{https://github.com/gd-zhang/Weight-Decay} without weight decay.

All models are trained for 200 epochs. The range of hyper-parameters is listed as below and some parameters we used in Section \ref{sec-5} are given in Table \ref{para}.

\begin{itemize}
  \setlength{\itemsep}{3pt}
  \setlength{\parsep}{0pt}
  \setlength{\parskip}{0pt}
  \item learning rate $\alpha$: \{1e-4, 3e-4, 1e-3, 3e-3, 1e-2, 3e-2, 1e-1, 3e-1, 1, 3\}. The learning rate in part \uppercase\expandafter{\romannumeral1} is kept constant. The learning rate in part \uppercase\expandafter{\romannumeral2} is tuned as follows. The initial learning rate is multiplied by 0.1 every 40 epochs for KFAC/TKFAC\_nor/TKFAC\_new and every 60 epochs for SGDM/Adam.
  \item damping $\lambda$: \{1e-4, 3e-4, 1e-3, 3e-3, 1e-2, 3e-2, 1e-1, 3e-1\} for KFAC/TKFAC\_nor; \{1e-8, 1e-4, 1e-3, 1e-2, 1e-1\} for Adam.
  \item the parameter to restrict trace $\nu$: \{1e-4, 1e-3, 1e-2, 1e-1, 1, 10, 100\} for TKFAC\_new.
  \item moving average parameter $\varepsilon$: 0.95 for KFAC/TKFAC\_nor/TKFAC\_new.
  \item momentum $\tau$: 0.9 for all methods.
  \item $T_{\rm{FIM}}=T_{\rm{INV}}=100$ for KFAC/TKFAC\_nor/TKFAC\_new.
  \item batch size: 500 for MNIST and 128 for CIFAR-10/CIFAR-100.
\end{itemize}

\begin{table}[htb]
\setlength{\abovecaptionskip}{0.0cm}
\setlength{\belowcaptionskip}{0.3cm}
\centering
\caption{Hyper-parameters used in Section \ref{sec-5} ($\alpha$ for SGDM, ($\alpha, \lambda$) for Adam/TKFAC\_nor/TKFAC\_new and ($\alpha, \lambda, \nu$) for TKFAC\_new).}
\label{para}
\resizebox{\textwidth}{!}{
\begin{tabular}{@{}c|c|ccccc@{}}
\toprule
                   &          & SGDM & Adam         & KFAC        & TKFAC\_nor       & TKFAC\_new         \\ \midrule
\multirow{2}{*}{part \uppercase\expandafter{\romannumeral1}} & FNN      & 0.01 & (1e-3, 1e-4) & (0.03, 0.03) & (0.03, 0.03) & -               \\
                   & VGG16    & 0.01 & (1e-3, 1e-8) & (1e-3, 1e-3) & (1e-3, 1e-3) & -               \\ \midrule
\multirow{2}{*}{part \uppercase\expandafter{\romannumeral2}} & VGG16    & 0.03  & (3e-4, 1e-8) & (1e-3, 1e-3) & (1e-3, 1e-3) & (1e-3, 1e-3, 1) \\
                   & ResNet20 & 0.03  & (1e-3, 1e-8) & (1e-3, 1e-3) & (1e-3, 1e-3) & (1e-3, 1e-3, 1e-2) \\ \bottomrule
\end{tabular}
}
\end{table}

\section{More results} \label{sec-b}
\subsection{Interpretation of the assumption in Section \ref{sec-4}}
\begin{figure}[H]
  \vspace{-0.35cm}
  \centering
  \includegraphics[width=0.65\linewidth]{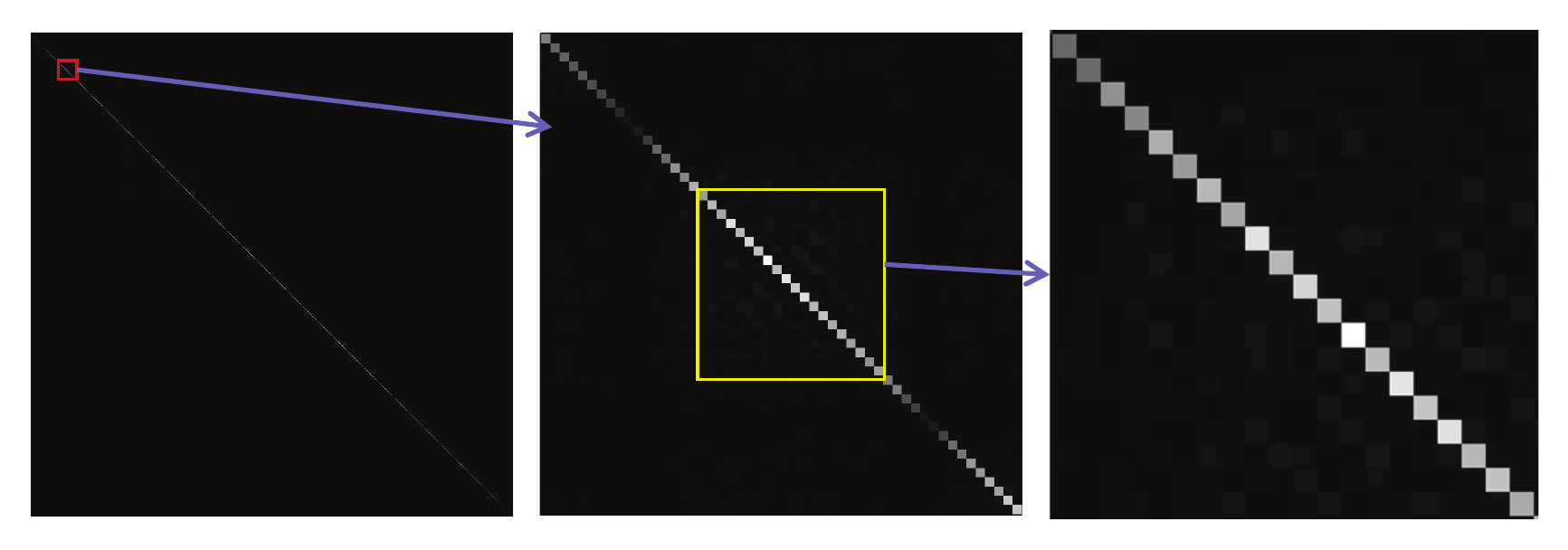}\\
  \caption{Visualization of the absolute values of the correlations between products at different spatial locations. We select a convolution layer from the Resnet32 network on the CIFAR-10 dataset using TKFAC. In this figure, the data is recorded in the initialization and we gradually enlarge the diagonal part of the figure.}\label{su-1}
\end{figure}

\begin{figure}[H]
	\centering
	\vspace{-0.35cm}
	\subfigtopskip=2pt
	\subfigbottomskip=2pt
	\subfigcapskip=2pt
	\subfigure{\includegraphics[width=0.2\linewidth]{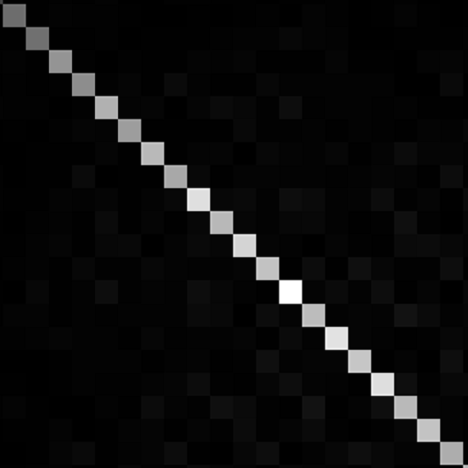}}
    \subfigure{\includegraphics[width=0.2\linewidth]{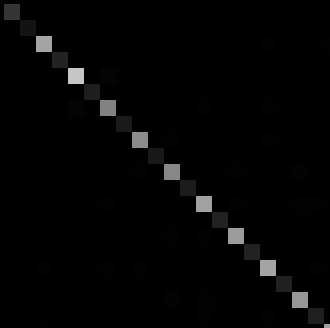}}
    \subfigure{\includegraphics[width=0.2\linewidth]{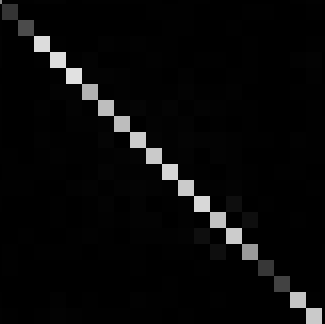}}\\
    \subfigure{\includegraphics[width=0.2\linewidth]{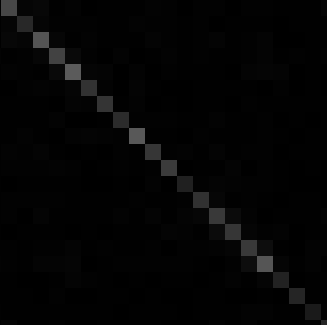}}
    \subfigure{\includegraphics[width=0.2\linewidth]{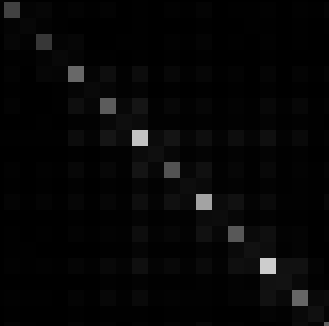}}
    \subfigure{\includegraphics[width=0.2\linewidth]{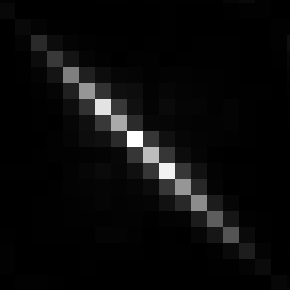}}\\
    \subfigure{\includegraphics[width=0.2\linewidth]{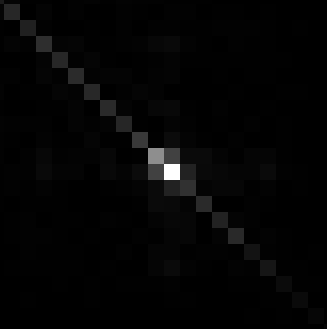}}
    \subfigure{\includegraphics[width=0.2\linewidth]{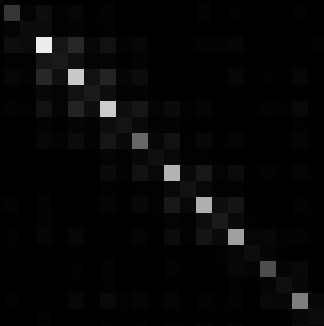}}
    \subfigure{\includegraphics[width=0.2\linewidth]{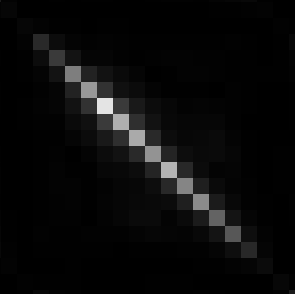}}
\caption{Visualization of the absolute values of the correlations between products at different spatial locations. We select three different convolution layers from the Resnet32 network on the CIFAR-10 dataset using TKFAC. The three figures in every column correspond to the same layer and the data is recorded in the initialization, 5000 iterations and 10000 iterations. We only show a part of the enlarged figures as Figure \ref{su-1}.}
	\label{su-2}
\end{figure}
In Figure \ref{su-1} and Figure \ref{su-2}, we show the correlations between products at different spatial locations. We show the data recorded in the initialization and gradually enlarge the diagonal part of the left figure in Figure \ref{su-1}. Figure \ref{su-2} shows the data recorded at different time for three different convolution layers. In the initialization (the first row in Figure \ref{su-2}), we can see that the products are very weakly correlated at two any different spatial locations. This means that our assumption is reasonable. What's more, this weak correlation can also be kept during training (the second and the third rows in Figure \ref{su-2}), which further proves the rationality of our assumption.

\subsection{Results of the test set in Part \uppercase\expandafter{\romannumeral1}}

\begin{figure}[htb]
	\centering
	\vspace{-0.35cm}
	\subfigtopskip=2pt
	\subfigbottomskip=2pt
	\subfigcapskip=2pt
	\subfigure[Testing error]{
		\includegraphics[width=0.3\linewidth]{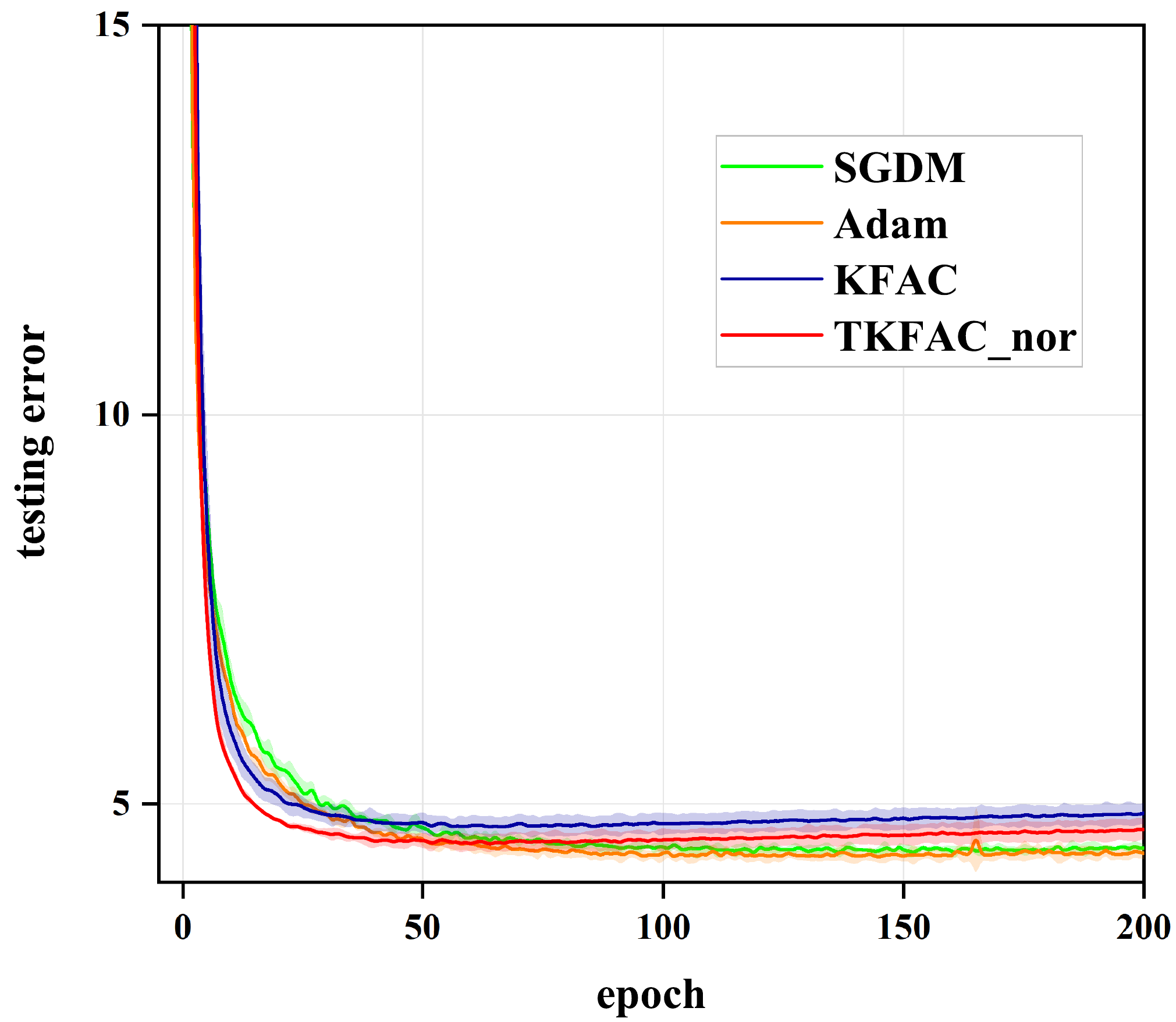}}
	\subfigure[Wall-clock time]{
		\includegraphics[width=0.3\linewidth]{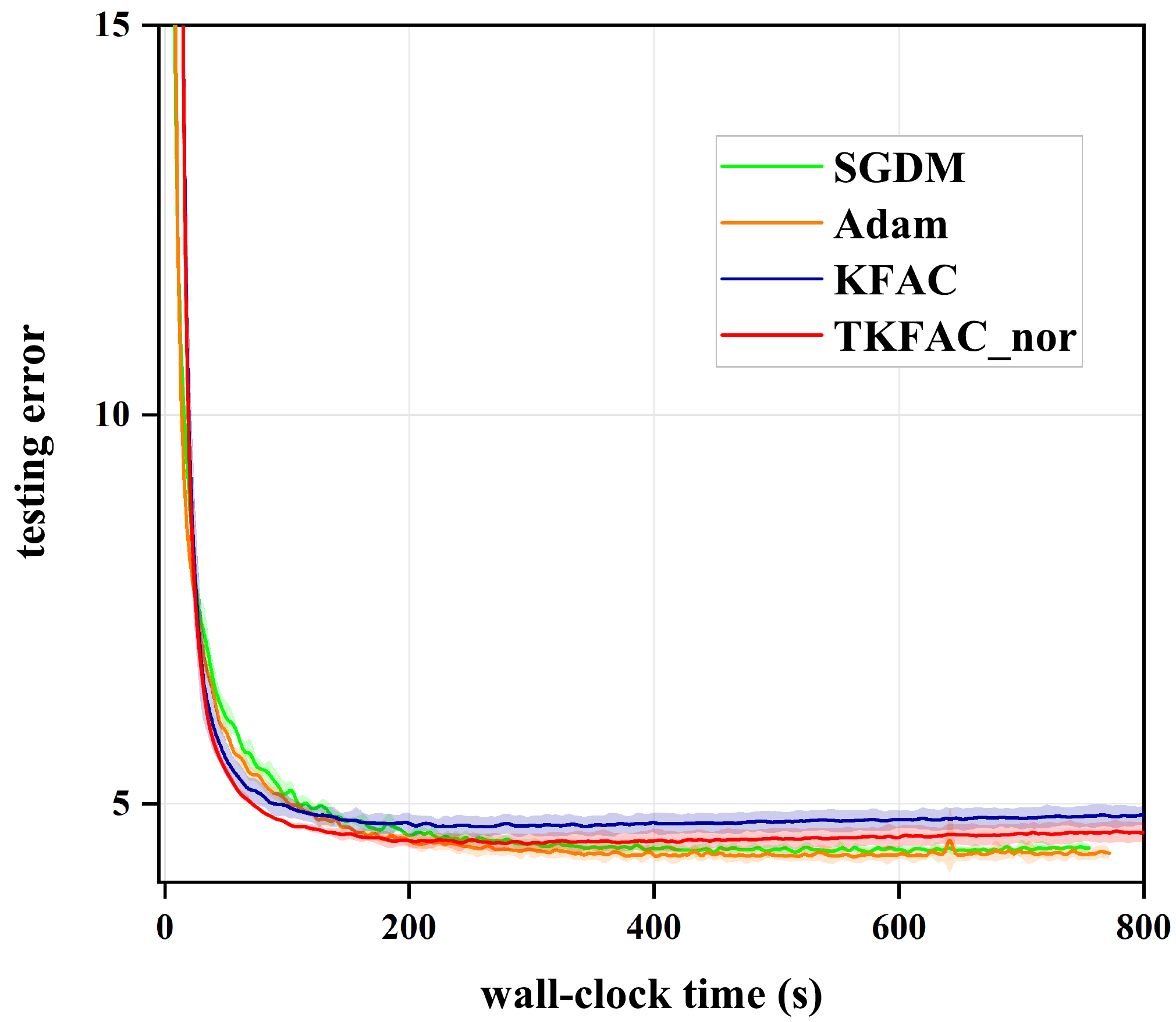}}
\caption{The curves of testing error on MNIST. (a) Error vs epochs; (b) Error vs wall-clock time.}
\end{figure}

\begin{figure}[htb]
	\centering
	\vspace{-0.35cm}
	\subfigtopskip=2pt
	\subfigbottomskip=2pt
	\subfigcapskip=2pt
	\subfigure[Testing accuracy]{
		\includegraphics[width=0.3\linewidth]{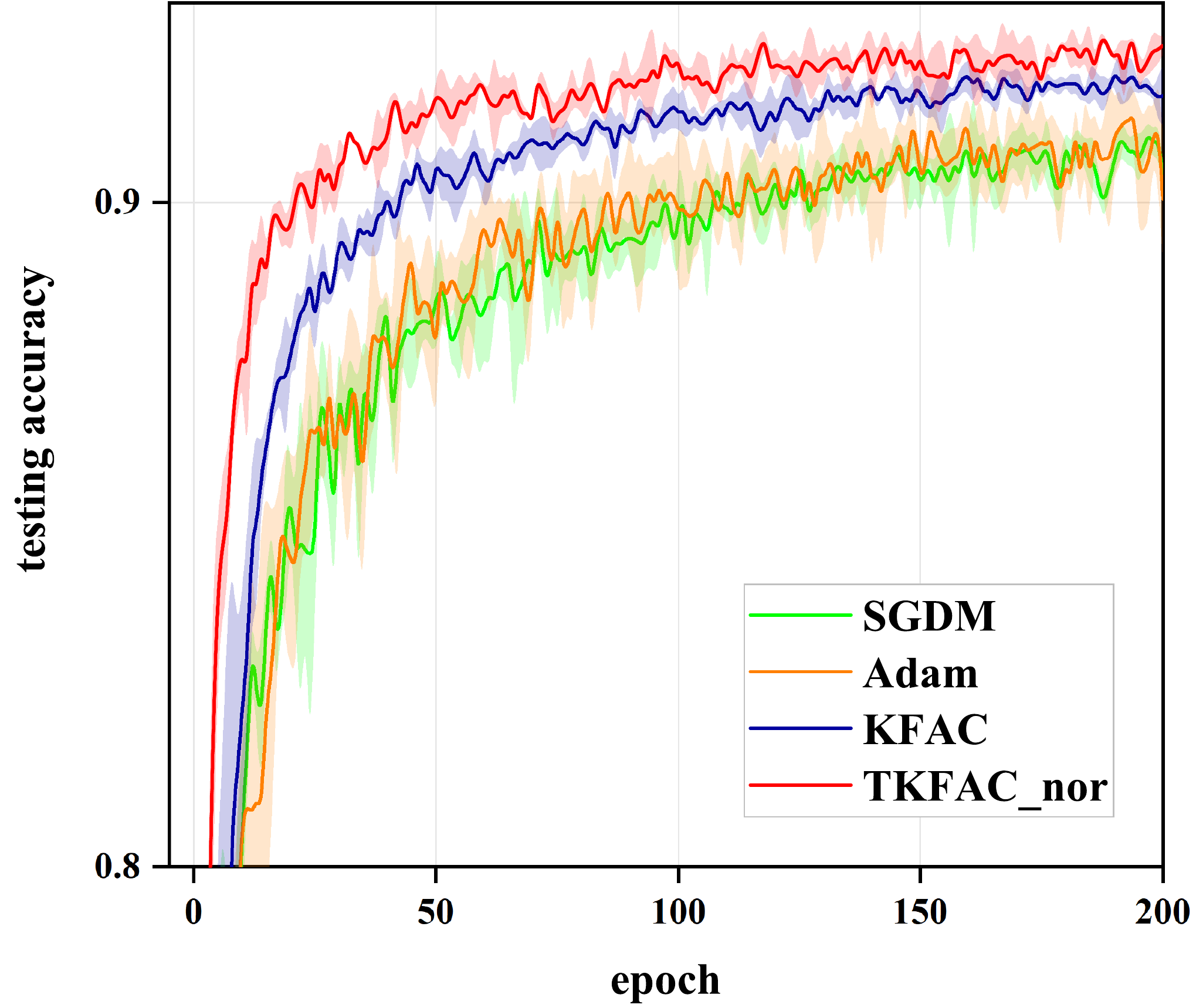}}
	\subfigure[Wall-clock time]{
		\includegraphics[width=0.3\linewidth]{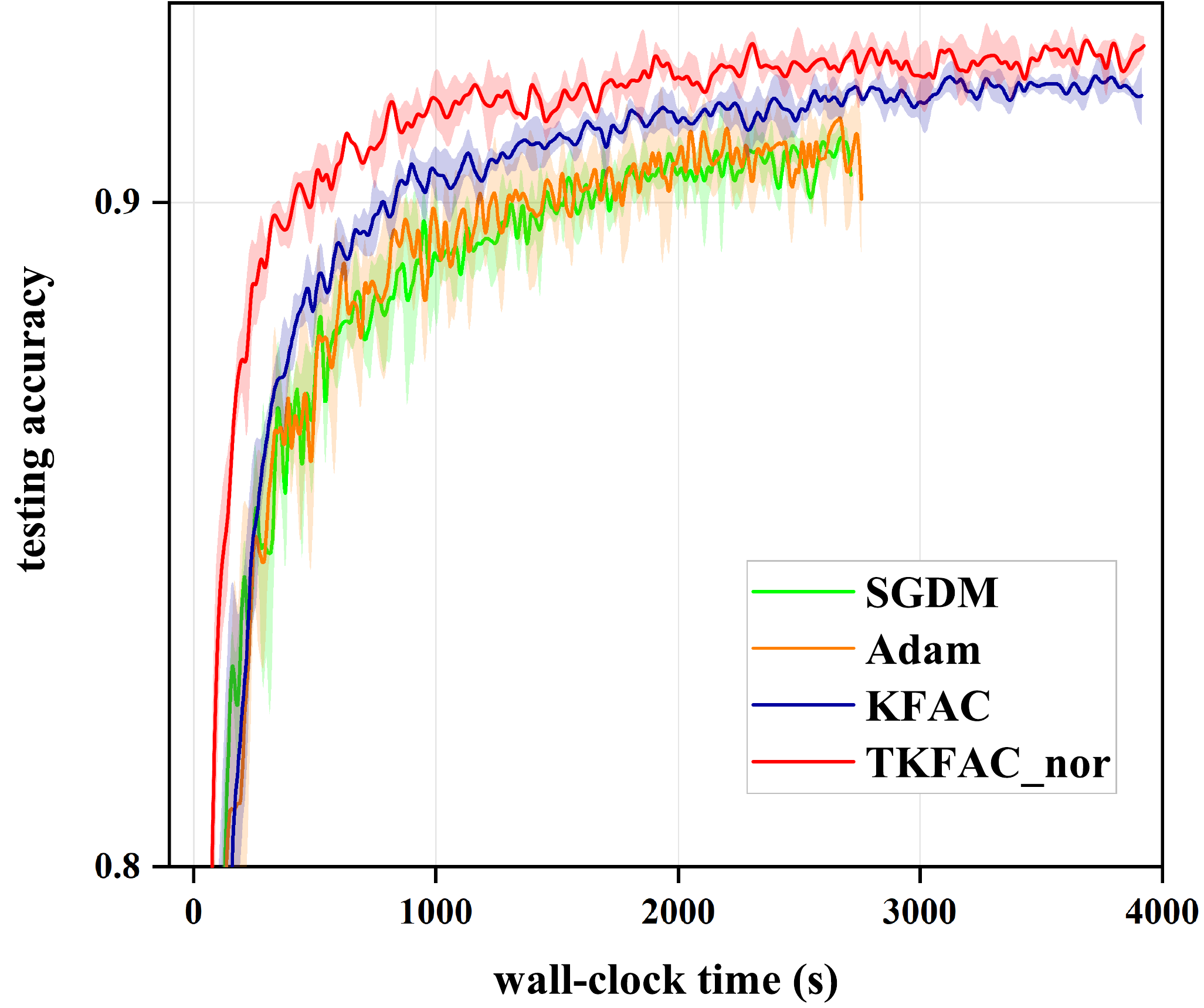}}
\caption{The curves of testing accuracy on VGG16 for CIFAR-10. (a) Testing accuracy vs epochs; (b) Testing accuracy vs wall-clock time.}
\end{figure}


\end{document}